\documentclass{article}
\usepackage{neurips_2024}
\usepackage[utf8]{inputenc} 
\usepackage[T1]{fontenc}    
\usepackage{hyperref}       
\usepackage{url}            
\usepackage{booktabs}       
\usepackage{amsfonts}       
\usepackage{nicefrac}       
\usepackage{microtype}      
\usepackage{xcolor}         
\usepackage{graphicx}
\usepackage{caption}
\usepackage{float}
\usepackage{booktabs} 
\usepackage{graphicx}
\usepackage{multirow}
\usepackage{pdflscape}
\usepackage{pifont} 
\usepackage{rotating} 
\usepackage{amsmath} 
\usepackage{algorithm}
\usepackage{algorithmic}
\usepackage{amsmath}
\usepackage{hyperref}
\usepackage{longtable}
\usepackage{enumitem}
\usepackage{wrapfig}
\usepackage{makecell}
\definecolor{c3}{HTML}{00bc12}
\newcommand{\cmark}{\textcolor{c3}{\ding{51}}}%
\newcommand{\xmark}{\textcolor{red}{\ding{55}}}%

\title{Chemistry3D: Robotic Interaction Benchmark for Chemistry Experiments}

\author{
\bf
{Shoujie Li}$^{1}$$^{\dagger}$,
{Yan Huang}$^{2}$$^{\dagger}$,
{Changqing Guo}$^{3}$$^{\dagger}$,
{Tong Wu}$^{1}$,
{Jiawei Zhang}$^{1}$,
~\\
\bf
{Linrui Zhang}$^{1}$,
{Wenbo Ding}$^{1}$$^{\#}$,
~\\
$1$ Tsinghua University, 
$2$ Wuhan University, 
$3$ South China University of Technology,
~\\
${\dagger}$ Equal Contribution
${\#}$ Corresponding author: ding.wenbo@sz.tsinghua.edu.cn.
}

\begin{document}

\maketitle
\begin{abstract}

The advent of simulation engines has revolutionized learning and operational efficiency for robots, offering cost-effective and swift pipelines. However, the lack of a universal simulation platform tailored for chemical scenarios impedes progress in robotic manipulation and visualization of reaction processes. Addressing this void, we present Chemistry3D, an innovative toolkit that integrates extensive chemical and robotic knowledge. Chemistry3D not only enables robots to perform chemical experiments but also provides real-time visualization of temperature, color, and pH changes during reactions. Built on the NVIDIA Omniverse platform, Chemistry3D offers interfaces for robot operation, visual inspection, and liquid flow control, facilitating the simulation of special objects such as liquids and transparent entities. Leveraging this toolkit, we have devised RL tasks, object detection, and robot operation scenarios. Additionally, to discern disparities between the rendering engine and the real world, we conducted transparent object detection experiments using Sim2Real, validating the toolkit's exceptional simulation performance. The source code is available at \href{https://github.com/huangyan28/Chemistry3D}{https://github.com/huangyan28/Chemistry3D}, and a related tutorial can be found at \href{https://www.omni-chemistry.com}{https://www.omni-chemistry.com}.


\end{abstract}

\section{Introduction}

Chemistry is a constantly evolving and experimental discipline\cite{leardi2009experimental}. The birth of a new substance or material often requires thousands of experiments. Therefore, chemical experiments are unfriendly to researchers. The tedious and repetitive nature of this work not only imposes immense labor intensity on researchers and chemical engineers (Some chemical engineers often work up to 50 hours a week in the US, says the Bureau of Labor Statistics\cite{engineer}) but also poses threats to their physical health due to exposure to harmful chemicals. In addition, many experiments consume large amounts of resources, with global consumption of chemicals calculated to exceed €5.77 trillion in 2022 alone\cite{consumption}.  In today's rapidly advancing era of embodied intelligence technology, proposing a 3D simulator that includes robot operations and chemical reaction processes is imperative, which not only improves the efficiency of experiments and reduces the cost of experiments but also liberates human beings from heavy scientific experimental tasks.


Chemical experiments contain many chemical manipulation and visual detection tasks, which are dangerous if the robot is trained directly in a real environment.  Although considerable research has been conducted into robot simulation systems\cite{van1990computer,dimian2014integrated,sun2018pyscf,motard1975steady,yuan2022pre}, a dedicated chemical 3D simulation system for robots has yet to be proposed.
 Current research on chemical robots mainly focuses on algorithmic aspects, such as organic synthesis methods\cite{granda2018controlling,de2019synthetic,lodewyk2012computational} and reinforcement learning(RL)\cite{beeler2023chemgymrl,rajak2021autonomous,zhang2021deep,sridharan2024deep,zhou2017optimizing} to improve yield. This is because chemistry and robotics are interdisciplinary fields, and designing a chemical 3D simulation system tailored for robots requires addressing numerous challenges, including but not limited to:
 \textbf{ (1) Immature rendering engines for liquids and transparent objects:} Chemical experiments involve many liquids and transparent objects, and achieving efficient and realistic rendering engines is difficult\cite{jiang2023robotic}.
\textbf{ (2)  Vast chemical reaction databases:} Chemistry is a complex discipline covering various reaction types such as organic, inorganic, liquid, solid, and gas; thus, implementing chemical simulation requires extensive database support\cite{gasteiger1990models}.
 \textbf{ (3) Complex calculation methods and parameters:} Real chemical reactions involve multiple parameters such as heat, temperature, pH, color, etc.; visualizing these parameters requires a deep understanding of chemistry and complex calculation methods\cite{engkvist2018computational}.
Therefore, to realize a chemical 3D simulation system tailored for robots, it is necessary to span multiple disciplines, such as chemistry, computer graphics, and robotics, and address the various technical challenges mentioned above.

\begin{figure}[H]
    \centering
    \includegraphics[width=1\textwidth]{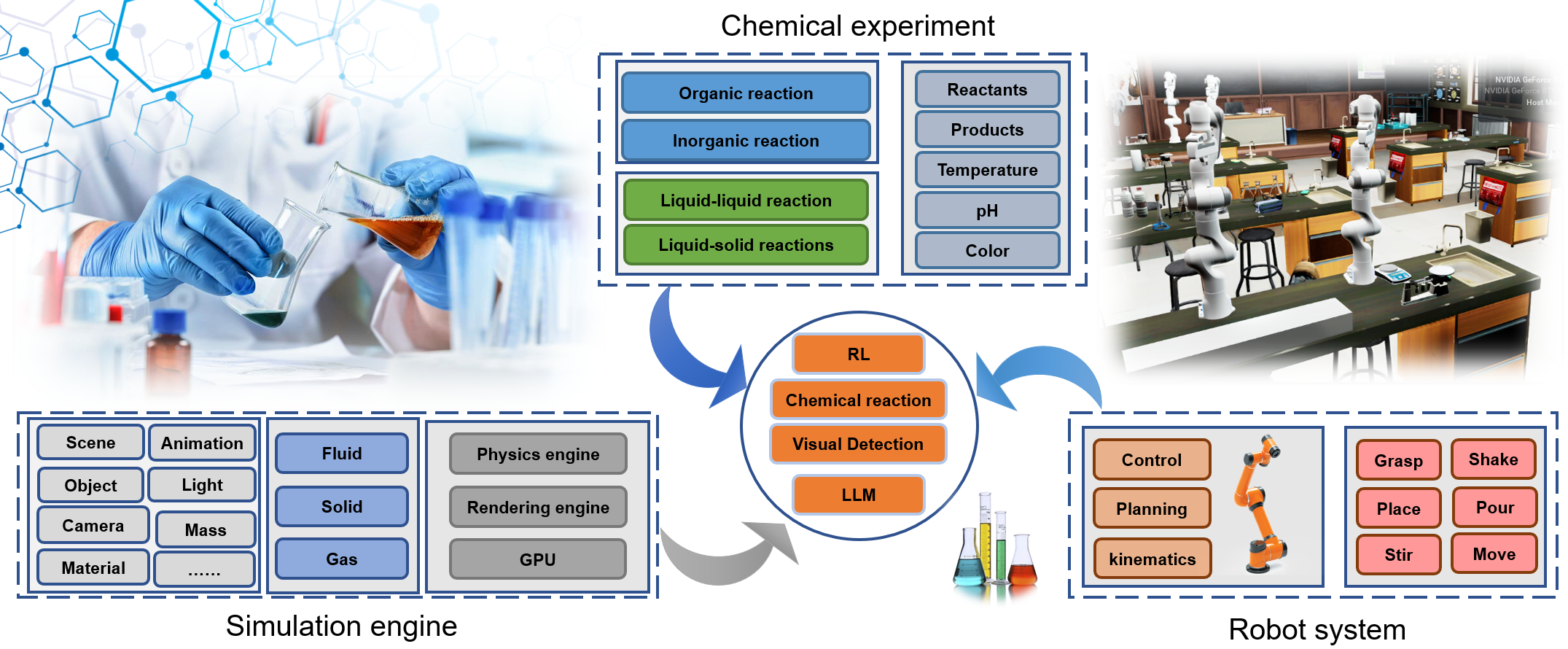}
    \caption{\footnotesize{Illustration of the Chemistry3D.  Chemistry3D integrates a robot system, chemical experiment, and simulation engine, providing interfaces for robot manipulation, visual inspection, and fluid flow control and enabling reaction visualization.}}
    \label{fig:Chemistry3D}
\end{figure}

The advancements in 3D rendering technology and the development of large language models (LLM) have presented us with new opportunities. Omniverse\cite{omniverse}, introduced by NVIDIA, is an open virtual collaboration and simulation platform encompassing a wide array of 3D modeling tools, renderers, animation tools, and physics engines. It enables robots to create more realistic and interactive virtual environments\cite{hummel2019leveraging}. Therefore, leveraging NVIDIA's Omniverse simulator, we propose a high-performance simulating toolkit for chemical experiments named Chemistry3D, as shown in Fig.\ref{fig:Chemistry3D}. This toolkit allows robots to conduct organic, inorganic, and various other experiments within 3D environments. Furthermore, to enhance the versatility of the simulator, we have opened convenient data interfaces, enabling operators to add unknown chemical reactions to the database effortlessly.
The contributions of this paper are as follows:
 \textbf{ (1) Novel 3D scene for chemical experiments:} We introduce a pioneering 3D scene tailored for chemical experiments. This scene not only features a plethora of chemical containers and robots but also facilitates functions such as transparent object simulation and fluid simulation.
 \textbf{  (2)  Establishment of a comprehensive chemical dataset:} We have curated a chemical dataset comprising over 1,000 inorganic reactions and 100,000 organic reactions. This dataset not only yields intermediate products for chemical simulation but also provides real-time feedback on changes in temperature, color, and pH during the experimental process.
  \textbf{  (3)   Performance validation through various experiments:} To validate the performance of our simulation, we conducted experiments including object grasping training based on RL, chemical experiment operations guided by LLM, and transparent object detection based on Sim2Real techniques. These experiments demonstrate the significant potential applications of our scene in the field of machine learning.

\section{Related Work}

In addressing the challenges of chemical reaction visualization and operation in the simulation environment, a variety of specialized tools and research efforts have been established. We conducted a comparative analysis of Chemistry3D and other tools in the domains of robotics and chemistry, as is summarized in Table \ref{tab:compare}. Notably, benchmarking robotic manipulation aspects of chemical experiments in simulation has not been previously addressed. Existing efforts typically focus on chemical reaction generation or robotic operations within specific scenarios. Specifically, our proposed work focuses more deeply on robot manipulation and embodied intelligence.

Traditionally, chemical reaction simulators primarily emphasize the molecular level. Tools from Interactive Chemistry\cite{ic} simulate molecular collisions, while techniques based on Computational Fluid Dynamics (CFD)\cite{cfd} could be used to simulate gas reactions. Combining these with data-driven techniques, MoleculeNet\cite{molecule} offers a large-scale benchmarking platform for molecular machine learning, proposing methods for feature characterization. Additionally, numerous databases in organic chemistry, such as ORD\cite{ord} and ORDerly\cite{orderly}, focus on structured reaction characterization. ChemSpider\cite{chemspider} serves as a chemical substance information retrieval tool. In inorganic chemistry, Chenaxon\cite{chemaxon} and RXN for Chemistry\cite{rxn,nature} are used for reaction prediction and synthesis pathway optimization. ChemReaX\cite{chemreax} provides examples based on a small sample database and includes information on thermochemistry and reaction intermediates.

Few simulations integrate chemical experiments with robotics. For operations, existing works tend to focus on specific tasks. For instance, Robot Air Hockey\cite{rah} is employed for Sim2Real applications in playing air hockey, while Panda MuJoCo Gym\cite{mujoco} benchmarks RL tasks such as pushing, sliding, and object manipulation. A Unity-based Simulator\cite{unity} creates game-like scenarios for experimental manipulation for educational purposes. For perceptions, CABD\cite{cabd} offers benchmark datasets for the image recognition of chemical apparatus. ChemGymRL\cite{beeler2023chemgymrl} provides detailed information on the intermediates of chemical reactions for RL. For system architecture design, different pipelines are developed for enabling autonomous laboratory, utilizing machine learning (ML)\cite{BENNETT2022100831} and natural language processing (NLP)\cite{chemos} methods. Furthermore, ARChemist\cite{archemist} designed several modular managers for processing the chemical recipe and interacting with the robots.

\begin{table}[h]
   \centering
    \caption{\small{Comparison of Chemistry3D with Other Researches and Tools}}
   \resizebox{\columnwidth}{!}{%
    \begin{tabular}{lcccccccccccccccccccccccc}
     & \rotatebox{60}{\makecell[c]{Chemistry3D\\ (ours)}} & \rotatebox{60}{\makecell[c]{Molecular\\Collision\cite{ic}}} & \rotatebox{60}{CFD\cite{cfd}} & \rotatebox{60}{MoleculeNet\cite{molecule}} & \rotatebox{60}{ORD\cite{ord}} & \rotatebox{60}{ORDerly\cite{orderly}} & \rotatebox{60}{ChemSpider\cite{chemspider}} & \rotatebox{60}{Chemaxon\cite{chemaxon}} & \rotatebox{60}{\makecell[c]{RXN for \\Chemistry\cite{rxn}}} & \rotatebox{60}{ChemReaX\cite{chemreax}} & \rotatebox{60}{\makecell[c]{Robot Air \\Hockey\cite{rah}}} & \rotatebox{60}{\makecell[c]{Panda MuJoCo\\Gym\cite{mujoco}}} & \rotatebox{60}{\makecell[c]{Unity-based \\Simulator\cite{unity}}} & \rotatebox{60}{CABD\cite{cabd}} & \rotatebox{60}{ChemGymRL\cite{beeler2023chemgymrl}} & \rotatebox{60}{\makecell[c]{ML-based \\architecture\cite{BENNETT2022100831}}} & \rotatebox{60}{ChemOS\cite{chemos}} & \rotatebox{60}{ARChemist\cite{archemist}} \\ \midrule
     \textbf{Manipulation Information} &  &  &  &  &  &  &  &  &  &  &  &  &  &  &  &  &  &  &   \\
     
     Perception Information         & \cmark & \xmark & \xmark & \xmark & \cmark & \cmark & \xmark & \xmark & \xmark & \xmark & \cmark & \cmark & \cmark & \cmark & \xmark & \xmark & \xmark & \xmark \\
     Transparent Object Detection   & \cmark & \xmark & \xmark & \xmark & \xmark & \xmark & \xmark & \xmark & \xmark & \xmark & \xmark & \xmark & \xmark & \cmark & \xmark & \xmark & \xmark & \xmark \\
     Contact Information            & \cmark & \xmark & \xmark & \xmark & \xmark & \xmark & \xmark & \xmark & \xmark & \xmark & \cmark & \cmark & \xmark & \xmark & \xmark & \xmark & \xmark & \xmark \\
     Reinforcement Learning         & \cmark & \xmark & \xmark & \xmark & \xmark & \xmark & \xmark & \cmark & \cmark & \cmark & \cmark & \cmark & \xmark & \cmark & \cmark & \cmark & \xmark & \xmark \\
     Embodied Intelligence          & \cmark & \xmark & \xmark & \xmark & \xmark & \xmark & \xmark & \xmark & \xmark & \xmark & \xmark & \xmark & \xmark & \xmark & \xmark & \xmark & \cmark & \xmark \\
     Task Subdivision               & \cmark & \xmark & \xmark & \xmark & \cmark & \cmark & \xmark & \cmark & \cmark & \xmark & \xmark & \cmark & \xmark & \xmark & \xmark & \xmark & \cmark & \cmark \\ 
     
     \midrule
     \textbf{Chemical Information} &  &  &  &  &  &  &  &  &  &  &  &  &  &  &  &  &  &  & \\
     Intermediate States            & \cmark & \xmark & \cmark & \xmark & \xmark & \xmark & \xmark & \xmark & \xmark & \cmark & \xmark & \xmark & \xmark & \xmark & \cmark & \xmark & \xmark & \xmark \\
     Color Information              & \cmark & \xmark & \xmark & \xmark & \cmark & \cmark & \cmark & \cmark & \xmark & \xmark & \xmark & \xmark & \cmark & \xmark & \cmark & \xmark & \cmark & \xmark \\
     Spectrum Information            & \cmark & \xmark & \xmark & \xmark & \xmark & \xmark & \cmark & \cmark & \xmark & \xmark & \xmark & \xmark & \xmark & \xmark & \cmark & \xmark & \xmark & \xmark \\
     Information Collection         & \cmark & \xmark & \xmark & \xmark & \xmark & \cmark & \cmark & \cmark & \xmark & \xmark & \xmark & \xmark & \xmark & \xmark & \xmark & \cmark & \cmark & \xmark \\
     Thermodynamic Property         & \cmark & \xmark & \xmark & \xmark & \cmark & \cmark & \cmark & \xmark & \cmark & \cmark & \xmark & \xmark & \xmark & \xmark & \xmark & \xmark & \xmark & \xmark \\
     pH Value                       & \cmark & \xmark & \xmark & \xmark & \xmark & \xmark & \xmark & \cmark & \xmark & \xmark & \xmark & \xmark & \xmark & \xmark & \xmark & \xmark & \cmark & \xmark \\
     Temperature                    & \cmark & \xmark & \cmark & \xmark & \cmark & \cmark & \xmark & \xmark & \xmark & \cmark & \xmark & \xmark & \xmark & \xmark & \cmark & \xmark & \xmark & \xmark \\
     Fluid Simulation               & \cmark & \xmark & \cmark & \xmark & \xmark & \xmark & \xmark & \xmark & \xmark & \xmark & \xmark & \xmark & \xmark & \xmark & \xmark & \xmark & \xmark & \xmark \\
     Database                       & \cmark & \xmark & \xmark & \cmark & \cmark & \cmark & \cmark & \cmark & \cmark & \cmark & \xmark & \xmark & \xmark & \xmark & \xmark & \xmark & \cmark & \xmark \\
     Organic Reaction               & \cmark & \xmark & \xmark & \xmark & \cmark & \cmark & \xmark & \cmark & \cmark & \xmark & \xmark & \xmark & \xmark & \xmark & \cmark & \cmark & \cmark & \cmark \\
     Inorganic Reaction             & \cmark & \cmark & \cmark & \xmark & \xmark & \xmark & \xmark & \cmark & \xmark & \cmark & \xmark & \xmark & \xmark & \xmark & \cmark & \cmark & \cmark & \cmark \\
     \bottomrule
    \end{tabular}%
   }
    \label{tab:compare}
  \end{table}

\section{Chemistry3D}
\label{headings1}

Chemistry3D encompasses three main aspects: chemistry simulation, virtual chemical laboratory environments, and robotic manipulation. Compared with previous chemistry benchmarks\cite{beeler2023chemgymrl}, Chemistry3D supports realistic scene simulations for visual inputs and enables physical interactions between robots and objects. In terms of chemistry simulation, it aims to provide accurate and detailed models of chemical reactions. Virtual chemical laboratory environments offer scenarios that closely mimic real chemical experiments, enhancing the authenticity and applicability of simulations. Regarding robotic manipulation, Chemistry3D facilitates the use of robots in performing and optimizing chemical reactions, aiming to seamlessly integrate robotic manipulation with chemical processes.

\subsection{Chemistry Simulation}

\textbf{Inorganic Reactions: }The simulation of inorganic reactions is designed based on a database encompassing both reaction and chemical substance information. The database includes data on 65 different chemical substances, each characterized by its color, enthalpy values, and physical state. Additionally, it contains information on 65 fundamental reactions, specifying the reactants, products, and their stoichiometric ratios. 

\begin{wrapfigure}{r}{0.5\textwidth} 
\includegraphics[width=0.5\textwidth]{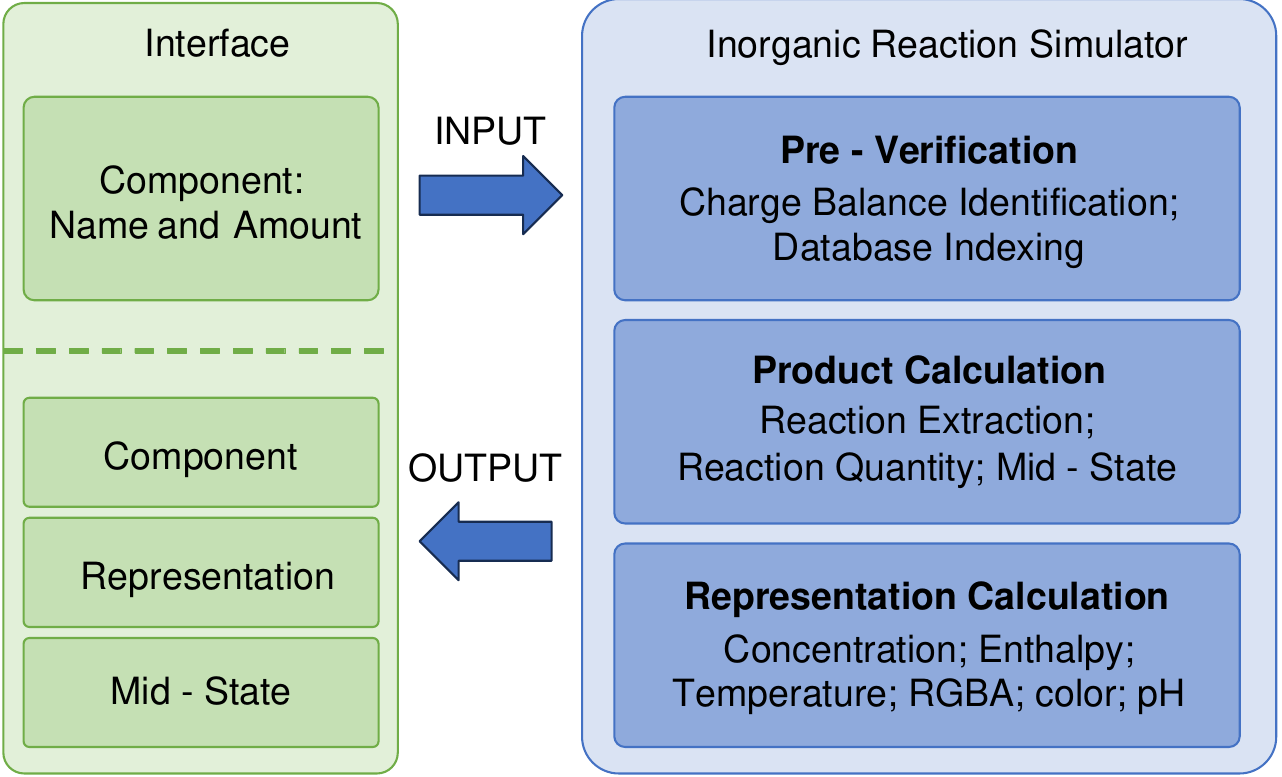}
\caption{\small{The framework of the inorganic reaction simulator. The simulator processes the input reactant components as a pipeline and output the product information on component, representation and mid-state.}}
\label{fig:sim}
\end{wrapfigure}

The simulator accepts input as a dictionary, with reactant names as keys and numbers of mole as values. The simulator is capable of performing iterative reactions, which allows for generating over 1000 possible reactions through various combinations.

For the output, the component interface outputs a dictionary of the same format, detailing the remaining reactants and reaction products. Furthermore, the representation interface provides a comprehensive output dictionary, including data on the color of the reaction mixture, enthalpy changes, pH levels, temperature changes, and the physical state of the resultant substances.



Precisely, the core framework of the inorganic reaction simulator is shown in Fig. \ref{fig:sim}. The detailed procedure within this simulator is discussed as follows.

\begin{itemize}[leftmargin=*]

    \item \textbf{Charge Balance Identification:} The initial step in the simulation process involves verifying that the input reactants satisfy charge conservation\cite{Giraud1978}. This ensures that the total charge of the reactants is balanced, which is crucial for confirming the set of reactants is realistic.

    \item \textbf{Database Indexing:} Following charge determination, the simulator verifies whether the identified reactants are present in the database by checking whether part of the current components can undergo a reaction.
    In this way, the simulator identifies the specific reactions by breaking down complex inorganic reactions into basic reactions\cite{keben}. The complex reaction is viewed as a sequence of these basic reactions executed in a specific order, determined by the sequential record of reactions in the database. For instance, acid-base reactions are premier to redox reactions. 

    \item \textbf{Reaction Extraction:} Once the inorganic reactions are identified by indexing the database, the simulator extracts the relevant reactants that are ready to react. The original input reactants is divided into two parts: the reacting one and the spectating one. By focusing on the specific ions involved, the simulator ensures that only the necessary reactions are considered in this reaction cycle, thereby streamlining the reaction prediction process.

    \item \textbf{Reaction Quantity Calculation:} After reaction extraction, the simulator calculates proportional amounts of reactants based on the stoichiometric coefficients. The simulator identifies which ion is completely consumed first, thereby establishing the limiting reagent. This allows the calculation of reaction quantity, i.e., how many moles of the "reaction equation" are involved in the process.
    Based on the input reagents, the simulator subtracts the amount of each reactant used and adds the generated products, yielding the final composition of substances. The stoichiometric calculations are vital for accurate reaction simulations and yield predictions.

    \item \textbf{Mid-State Calculation} The computation of intermediate states is based on the rate equation as Eqn.~\ref{rate}. This process involves considering the initial and final compositions of substances involved in the reaction to determine their temporal evolution\cite{Heald1974,doi:10.1021/j100324a007}. Specifically, by transforming the reaction order into unity, the simplified rate equation reveals an exponential decay relationship between substance compositions and time, resembling a negative exponential function with an offset. Consequently, by defining suitable reaction rate constants and utilizing the initial and final compositions of substances, temporal intermediate data can be obtained as a list.
    \begin{footnotesize} 

\begin{equation}
\label{rate}
    \text{rate} = k \cdot c(\text{A})^{m_a} \cdot c(\text{B})^{m_b}
\end{equation}
\end{footnotesize} 

    \item \textbf{Concentration Calculation:} The simulation proceeds by adding the total volume $V$ of the mixture, then calculating the concentration $c$ as Eqn. \ref{c}. This calculation forms the basis for subsequent concentration-dependent computations.
    \begin{footnotesize} 

\begin{equation}
\label{c}
        c = \frac{n}{V} 
\end{equation}
\end{footnotesize} 
    
    \item \textbf{Enthalpy Change Calculation:} The simulator then computes the enthalpy change based on reaction quantity as Eqn.~\ref{q1}. This step involves determining the heat absorbed or released during the reaction, which is essential for understanding the thermodynamics of the process. The total enthalpy change $Q$ is calculated by multiplying the reaction quantity $N$ to the enthalpy change per equation $\Delta H$, which is recorded in the database.
    \begin{footnotesize} 

\begin{equation}
\label{q1}
        Q = N \cdot \Delta H 
\end{equation}
\end{footnotesize}

    \item \textbf{Temperature Change Calculation:} Utilizing the specific heat capacity $C$ of the solvent, the simulator calculates the temperature change $\Delta T$ resulting from the reaction. The change is determined by Eqn.~\ref{q}, where $\rho$ is the density and $V$ is the volume of the solution. Temperature changes can influence reaction rates and equilibria, making this step important for dynamic analysis.
\begin{footnotesize} 

\begin{equation}
\label{q}
    \Delta T = \frac{Q}{C \cdot \rho \cdot V}
\end{equation}
   \end{footnotesize} 
 
    \item \textbf{RGBA Color Calculation:} The simulator assesses the color change in the reaction mixture by considering several factors:
    
    \begin{itemize}
        \item \textbf{Transparency Calculation:} It first calculates the transparency of the solution based on the concentration of the reactants and products, keeping the RGB values constant. The transparency value is determined through the exponential model derived by the dilution of solution\cite{doi:10.1021/acs.jced.5b00018,wei2006}, as Eqn.~\ref{a}. $c$ is the concentration of the reagent and $K$ is a constant representing the standard value for transparency.

        \begin{footnotesize} 

        \begin{equation}
        \label{a}
    a = 1 - 10^{-K \cdot c}
\end{equation}
\end{footnotesize} 

        \item \textbf{State Priority Consideration:} The simulator assigns priority to different physical states of the substances. If a substance exists in the solid state, it contributes to opacity, resulting in a turbid liquid. For substances in the liquid state, color mixing occurs based on RGBA values. However, the color of gaseous substances is not factored into the overall color calculation.
        
        \item \textbf{Color Mixing:} A negative mixing model is applied to determine the resultant color of the mixture of solution without solid substance\cite{Sundararajan2017}. This model helps in accurately simulating the combined color effects of multiple reactants and products.

        \item \textbf{Spectrum to RGB Transformation:} The spectrum information is included in this simulator, especially UV/Vis spectrum. The visible light spectrum of a substance determines its color appearance in a colorless transparent solution\cite{Udayakumar2014}. Therefore, the UV/Vis spectrum can be converted into RGB color values\cite{Fortner1997,cie} for visual display in simulation.
    \end{itemize}

    \item \textbf{pH Calculation:} Finally, the simulator calculates the pH of the solution through several sub-steps:

\begin{itemize}
    \item \textbf{Ionization Constant Interpolation:} Initially, a table is established to record water's ionization constant ($K_w$) in the liquid state at standard atmospheric pressure, ranging from 0 to 100 degrees Celsius. This table serves as a reference for interpolating $K_w$ values at any given temperature, facilitating precise pH calculations.

    \item \textbf{Electrolyte Ionization:} The simulator ignores the ionization of weak electrolytes and focuses on calculating the concentration of hydrogen ions ($H^+$) or hydroxide ions ($OH^-$) from strong electrolytes in water\cite{Burgot2017}. This step ensures that the major contributors to the pH are considered, as Eqn.~\ref{kw}.
    \begin{footnotesize} 

        \begin{equation}
        \label{kw}
    K_w = c(\text{H}^+) \cdot c(\text{OH}^-)
\end{equation}
    \end{footnotesize} 

    \item \textbf{pH Determination:} Based on the ionization constant and the concentrations of hydrogen or hydroxide ions, the simulator calculates the pH of the solution. Accurate pH determination is critical for understanding the acidity or basicity of the reaction environment.
    \begin{footnotesize} 

    \begin{equation}
    \text{pH} = -\log c(\text{H}^+)
\end{equation}
\end{footnotesize} 
\end{itemize}
\end{itemize}

\textbf{Organic Reactions: }The simulation of organic reactions integrates data from RXN for Chemistry\cite{rxn} for reaction information and ChemSpider\cite{chemspider} for chemical substance information. 
Similar to the inorganic simulator, this simulator also has a component and representation interface as output. The system accepts reactants represented by SMILES\cite{smile} strings and outputs the corresponding products for reaction product prediction. It’s also capable of predicting reaction yields, therefore determines the component output for the specific reaction.
For the representation, the simulator employs web scraping to query substance information; given a SMILES string, it retrieves a dictionary of properties and values. For more comprehensive data, users can query additional information using the CAS number\cite{cas} of the substance.

\textbf{Simulator Interface: }To integrate the chemical aspects with the operational aspects seamlessly, the simulator is embedded into a container class, which features three main methods: initialization, updating, and information retrieval. 
The initialization method distinguishes between organic and inorganic reactions and sets the chemical components by name, amount, and volume.
The updating method simulates sampling or mixing operations and can automatically conduct reactions. The output of this method represents the intermediate state of the container's contents, calculated through rate equations with an adjustable time step to suit simulation tasks in Omniverse. 
The information retrieval method allows access to component or representation information, enabling direct queries about concentration, color, and other properties for any container. 
This approach binds chemical information to simulated reagent bottles, facilitating clear demonstrations. It also aligns chemical reactions with operational actions, making the simulation intuitive. 
This integrated simulating method allows for accurate predictions and detailed representation calculations. It's essential for further studies in chemistry, including analysis of intermediate states and RL.

\begin{figure}[H]
    \centering
    \includegraphics[width=0.9\textwidth]{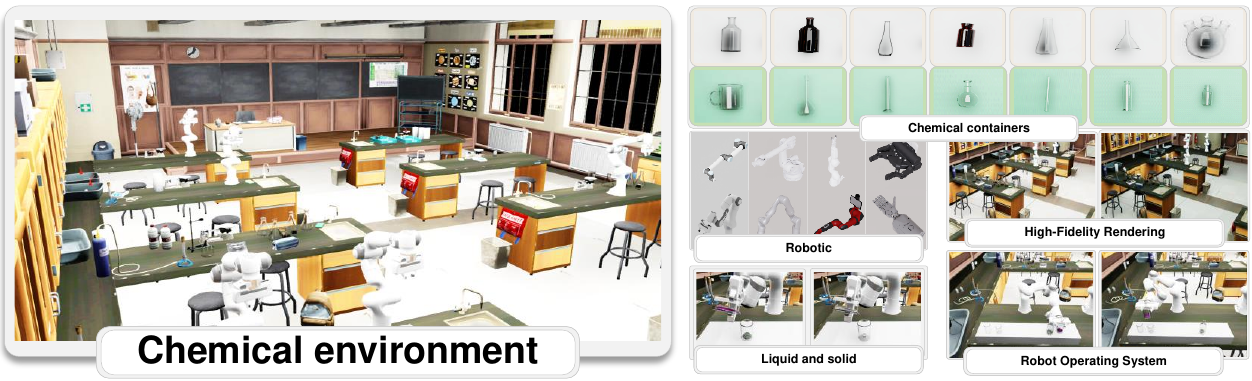}
    \caption{\small{Illustration of simulated chemistry environment. The chemical containers with a green background were 3D scanned from real objects. This environment is designed to support a wide range of chemical and robotic experiments, providing a highly detailed and interactive platform..}}
    \label{fig:Chemistry_scene}
\end{figure}

\subsection{Chemical Environment}
Chemistry3D offers an advanced and meticulously designed environment for simulating chemical experiments, as shown in Fig.~\ref{fig:Chemistry_scene}. This environment, built upon the NVIDIA Omniverse platform, integrates a variety of features essential for both chemical and robotic research.

\textbf{Rich Chemical Assets:}
Chemistry3D is endowed with an extensive series of chemical containers and instruments, meticulously designed to facilitate a diverse range of chemical reactions. These encompass both organic and inorganic reactions, as well as liquid-liquid and liquid-solid interactions. This vast collection of chemical assets allows for the simulation of various chemical experiments, offering researchers and educators a versatile platform to explore different reaction dynamics.

\textbf{Robotic Assets:}
The environment is equipped with numerous robotic arms and robotic grippers, enhancing the potential for robotic simulations. These robotic systems are capable of performing precise tasks such as grasping, shaking, pouring, stirring, placing, and moving chemical containers. This capability significantly expands the possibilities for robotic experimentation and automation within the chemical laboratory setting.


\textbf{Fluid and Rigid Body Simulation:}
Chemistry3D excels in simulating both fluid and rigid body interactions with high fidelity. This capability is particularly beneficial for visualizing intricate chemical processes involved in both inorganic and organic experiments. For instance, the platform can realistically simulate the dissolution of solid compounds in a liquid. Additionally, it can accurately depict the merging of two liquids and the resultant color changes, providing a vivid representation of reaction progress and intermediate states.

\textbf{High-Fidelity Rendering:}
The platform supports highly realistic rendering and light simulation, which is essential for accurately representing chemical experiments involving transparent materials. In many chemical lab settings, instruments such as glass beakers and flasks are transparent, posing challenges for vision-based tasks. Chemistry3D excels in simulating these transparent objects, providing detailed visual representations of chemical reactions, including changes in color and clarity. This enhances the effectiveness of visual inspections and analyses, which is crucial for monitoring and understanding chemical processes.

\textbf{Robot Operating System:}
Chemistry3D inherits robust support from Isaac-Sim, including integration with ROS and ROS2. This compatibility allows for a diverse range of robotic development and experimentation, enabling users to leverage advanced robotic operating systems for controlling and simulating robotic behavior within the chemical environment. This integration is crucial for developing and testing sophisticated robotic applications and workflows.

\subsection{Robotic Manipulation}

Chemistry3D aims to integrate robotic operations with the simulation of chemical experiments. Chemistry3D is built on the Nvidia Omniverse and PhysX 5 platforms, with robotic manipulations implemented using IsaacSim. This platform enables realistic simulations of rigid bodies, fluids, soft bodies, and other materials, thereby enhancing the overall realism of Chemistry3D through accurate physics simulation and light rendering effects.

Most chemical manipulations involve motion states such as grasping, moving, and pouring. Consequently, Chemistry3D proposes three distinct tasks, chemical experimental manipulation, embodied intelligence manipulation, and RL tasks, to advance the development of robotic operations within Chemistry3D. The descriptions of these tasks are as follows:

\textbf{Chemistry manipulation:}
Chemical experimental manipulations frequently involve the desired motions of target objects, e.g., pouring. Consequently, the tasks in Chemistry3D encompass a variety of chemical experimental operations. We have selected four common chemical experimental operations including picking, placing, pouring, stirring, and shaking. In our experiments, we developed simulation tasks within Chemistry3D to demonstrate these operations.

\textbf{Embodied intelligence:}
Embodied intelligence involves the interaction of semantic information between agents and humans, enabling robots to understand and perform desired chemical operations. This capability is vital for the automation of chemical processes. We designed specific chemical experiment scenarios to demonstrate that the development of embodied intelligence is possible in Chemistry3D. These scenarios showcase the potential for robots to autonomously observe the environment and complete specified tasks within the platform.

\textbf{RL task:}
RL has recently achieved success across a wide range of continuous control problems. Chemical experimental manipulations are often continuous processes, making RL an essential tool for integrating chemistry experiments with robotic simulation. IsaacGym is a well-regarded simulation environment specifically designed to support robot learning. The Omniverse Isaac Gym Reinforcement Learning (OmniIsaacGymEnvs) for Isaac Sim repository provides RL examples compatible with IsaacSim and has become a widely adopted environment for RL research. Utilizing this repository, we implemented a reward function setup similar to the provided examples and successfully achieved the picking of chemical containers. The RL task demonstrates that Chemistry3D can support RL research in robotic manipulation.




\begin{figure}[H]
    \centering
    \includegraphics[width=0.85\textwidth]{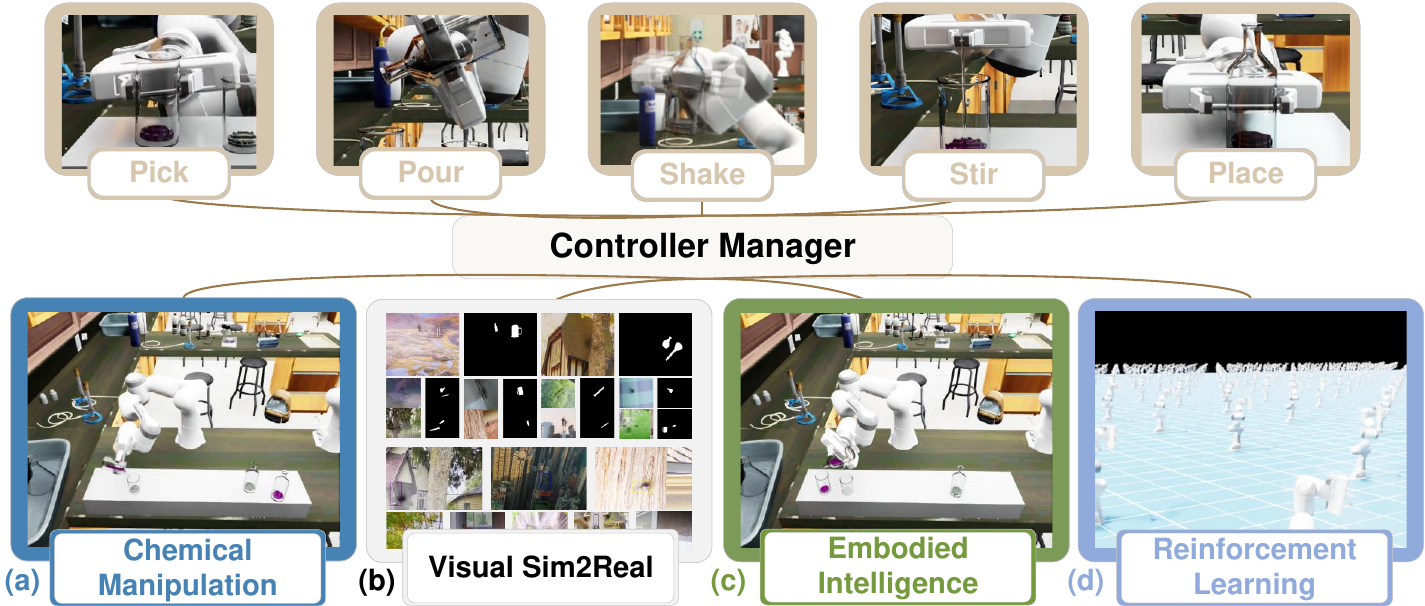}
    \caption{\small{Illustration of the motion and experiments in Chemistry3D. The figure showcases stages of Pick, Pour, Shake, Stir, and Place motions. Our experimental workflow integrates Chemical Manipulation, Visual Sim2Real, Embodied Intelligence, and Reinforcement Learning.}}
    \label{fig:Experiments}
\end{figure}

\section{Experiments}
\label{headings}

In this section, we investigate the capabilities of chemical reaction simulation and robotic manipulation within the Chemistry3D. All experiments are conducted within a simulated chemistry environment. As shown in Fig.~\ref{fig:Experiments}, we have carried out four experiments based on Chemistry3D, covering chemical manipulation, visual Sim2Real, embodied intelligence, and RL. Additionally, we have integrated the chemical simulator to perform both organic and inorganic experiments.


\subsection{Chemical Experiments}
\label{chemical_experiment}

In chemical experiments, we focus on inorganic and organic reactions, integrating robotic operations within IsaacSim to enhance the experimental process. (See Supplementary Material for more details.)

\textbf{Inorganic Experiment:} We selected redox reactions due to their notable color and state changes. Experiments involved potassium permanganate (KMnO\textsubscript{4}) with ferrous chloride (FeCl\textsubscript{2}), and hydrochloric acid (HCl) with iron(II) oxide (FeO). The center of mass of reactants determines contact points, triggering color and state transformations. Chemistry3D outputs detailed reaction data such as temperature, enthalpy change, and pH at each time step.

\textbf{Organic Experiment:} Focused on the simulation of mid-state products rather than color changes, we synchronized reaction steps with IsaacSim simulation. Using the reaction between Br\textsubscript{2} and C\textsubscript{20}H\textsubscript{12} as an example, mid-state products are generated upon reactant contact. This enables real-time optimization of final products and robotic manipulations.

\subsection{Chemical Manipulation}
As shown in Fig.~\ref{fig:Experiments}(a), chemical manipulation often involves tasks such as picking and placing. Our experiments demonstrated the effective deployment of robotic picking and pouring operations in chemical processes. We designed modular operations, including picking, pouring, shaking, stirring, and placing, managed by a Controller Manager. This manager ensures the sequential and integrated execution of these operations. We successfully combined picking, pouring, and placing operations, as illustrated in our experimental results. (See Supplementary Material for more details.)

\subsection{Visual Sim2Real}

To demonstrate Omniverse's capability in implementing an effective Sim2Real system, we selected complex transparent chemical containers for our experiments, as shown in Fig.~\ref{fig:Experiments}(b). We introduced 3D-scanned transparent models that are common in chemical labs. These models were used to test Sim2Real transfer for semantic segmentation and object detection tasks. We presented two simulation datasets and integrated Segmentation Models PyTorch\cite{Iakubovskii} for algorithm comparisons.(See Supplementary Material for more details.)

High-fidelity visualization in Omniverse enabled effective training of encoder-decoder networks, maintaining robust segmentation in real-world scenarios. We selected several mainstream encoder-decoder combinations for quantitative experiments. As illustrated in Table~\ref{tab:performance}, quantitative comparisons using Intersection over Union (IoU), Pixel Accuracy (PA), F1-Score, and F2-Score showed that the EfficientNet-DeepLabV3 combination outperformed others, achieving top scores across all metrics. 
 \begin{footnotesize} 
\begin{table}[H]
    \centering
    \caption{Performance Metrics of Different Encoder-Decoder Combinations}
    \label{tab:performance}
    \begin{tabular}{cccccc}
        \toprule
        Encoder & Decoder & IoU & PA & F1-Score & F2-Score \\
        \midrule
        TGCNN\cite{tgcnn} & - & 0.6604 & 0.9875 & 0.7851 & 0.7220 \\
        ResNet\cite{resnet} & Unet\cite{unet} & 0.7097 & 0.9902 & 0.8181 & 0.7813 \\
        VGG\cite{VGG} & Unet++\cite{unet++} & 0.6963 & 0.9896 & 0.8056 & 0.7494 \\
        EfficientNet\cite{efficientnet} & DeepLabV3\cite{DeepLabV3} & 0.7582 & 0.9917 & 0.8558 & 0.8112 \\
        \bottomrule
    \end{tabular}
    \captionsetup{justification=centering}
    
\end{table}
 \end{footnotesize} 
Additionally, we further validated the Sim2Real ability by performing an object detection task. For this task, we selected YOLO\cite{yolo} as our algorithm. The model trained within the simulation environment was evaluated for object detection in both simulated and real-world environments. The results are consistent with results in the semantic segmentation task, confirming that Chemistry3D effectively supports visual Sim2Real.

\subsection{Embodied Intelligence}

To evaluate Chemistry3D in the context of Embodied AI, we designed a scene replicating an inorganic chemistry experiment as shown in Fig.~\ref{fig:Experiments}(c). This setup included containers with KMnO\textsubscript{4}, FeCl\textsubscript{2}, and empty beakers. We developed agents for robotic manipulations, allowing the robot to observe, predict potential chemistry reactions, and execute tasks via natural language commands.

The robot acquires environmental data by accessing the positional and labeling information of objects. Based on its chemical knowledge base from Chemistry3D, the robot can predict potential chemical reactions. Upon receiving task directives from a human operator, it utilizes a Large Language Model (LLM) to generate and strategically plan the necessary operations, thereby ensuring the stability and accuracy of experimental procedures. (See Supplementary Material for more details.)

\subsection{Reinforcement Learning}

OmniIsaacGymEnvs facilitates complex RL tasks in Chemistry3D. We demonstrated the capability of RL research by setting the RL task of picking as shown in Fig.~\ref{fig:Experiments}(d). Using Proximal Policy Optimization (PPO)\cite{PPO} as the algorithm, The experiment involved 2048 environments, 3500 epochs, and a learning rate of \(5 \times 10^{-4}\), applied consistently across multiple experiments. We plotted the reward and success rate curves, showing robust outcomes. The results confirmed that robotic arms could successfully grasp chemical containers within Chemistry3D. (See Supplementary Material for more details.)

\section{Conclusion}

We presented a 3D robot simulation toolkit based on NVIDIA's Omniverse platform for chemical experiments. This system encompasses various chemical containers and robotic models, supporting transparent objects and fluid simulations. We have established an extensive chemical dataset that provides real-time feedback on various parameter changes during experiments. Through RL tasks, large language modeling, and Sim2Real experiments, we have demonstrated the significant potential of this system in machine learning applications. This system enhances the visualization and interactivity of chemical experiments and offers a new tool for interdisciplinary research in chemistry and robotics, promising to advance related fields.

\clearpage

\section{Supplementary Material}

\subsection{Mathematical Principle of Inorganic Reaction Simulator}

This section reveals the mathematical principles in the inorganic reaction simulator. For clear mathematical representation, first we define a single reagent with its amount as base unit $\mathbf{r}$. The input reactant is defined as a set $\mathbf{R}$, which satisfies:
\begin{footnotesize} 
\begin{equation}
    \mathbf{R = \{r_1,...,r_i\}}.  
\end{equation}
\end{footnotesize}

An ionic chemical reaction is defined as Eqn.\ref{react}, where $\mathbf{R}$ represents the set of reactants and $\mathbf{P}$ represents the set of products. Noted that if the reaction solvent is water, the reagent should be expressed as its ionized results. In this way, the reaction $\mathbf{C}$ contains component and stoichiometric information of a chemical reaction.
\begin{footnotesize} 
\begin{equation}
    \mathbf{C=[R,P]}
    \label{react}
\end{equation}
\end{footnotesize}

The complex reaction $\mathbf{C_k}$ is viewed as a sequence of these basic reactions $\mathbf{C_{k_1},...,C_{k_n},}$ executed in a specific order, determined by the sequential record of reactions in the database. The composition of reactions should obey the following equation:
\begin{footnotesize} 

\begin{equation}
    \mathbf{C_k = [\bigcup_{i=1}^{n}C_{k_i}[0],\bigcup_{i=1}^{n}C_{k_i}[1]] \equiv [\bigcup_{i=1}^{n}R_{k_i},\bigcup_{i=1}^{n}P_{k_i}]}
\end{equation}
\end{footnotesize} 

The database of ionic chemical reactions is constructed as a set $\mathbf{D_C}$ including numerous reactions $\mathbf{C_i}$, which satisfies:
\begin{footnotesize} 

\begin{equation}
    \mathbf{D = \{C_1,...,C_i}\}
\end{equation}
\end{footnotesize}

After the simulator accepts the input reactant, it checks whether part of the current components can undergo a reaction $\mathbf{C_m}$, i.e., if they are present in the database, which satisfies:
\begin{footnotesize} 

\begin{equation}
    \mathbf{C_m[0] \equiv R_m \subseteq R_{input}}
\end{equation}
\end{footnotesize} 

Therefore, the data structure used for database indexing must process entries sequentially from the beginning. This step is essential to ensure that all reactants and products are recognized and that their properties are well-defined within the system.

To start reacting, the original input reactants $\mathbf{R}$ is divided into two parts: the reacting one $\mathbf{\hat{R}}$ and the spectating one $\mathbf{\tilde{R}}$. These two parts can be expressed as:

\begin{footnotesize} 

\begin{equation}
    \mathbf{R_{input}} = \mathbf{\hat{R} \cup \tilde{R}},\mathbf{\hat{R} \cap \tilde{R}} = \emptyset
\end{equation}
\end{footnotesize} 

After extracting the relevant reactions, the simulator calculates proportional amounts of reactants $\mathbf{\bar{R}}$ based on the stoichiometric coefficients. The reaction quantity $N$ is determined as the minimum stoichiometric coefficients in proportional amounts of reactants. The calculation should satisfy:
    
\begin{footnotesize} 

\begin{equation}
    \mathbf{\hat{R}\stackrel{stoichometry}{\longrightarrow}\bar{R}}, \mathbf{N = min\{\bar{R}\}}
\end{equation}
\end{footnotesize} 

Finally, the simulator subtracts the consumed amount of each reactant from input reagents and adds the generated products, yielding the final composition of substances $\mathbf{R_{new}}$. The updated component set is calculated as follows:
\begin{footnotesize} 

\begin{equation}
    \mathbf{R_{new} = R_{input} - N (C_m[0] - C_m[1])}  
\end{equation}
\end{footnotesize}

\subsection{Inorganic Reaction Database}

The database of the inorganic reaction simulator is primarily divided into two parts: inorganic reaction information and chemical substance information. The inorganic reaction information refers to the data for ionic reactions, defined by their reactants, products, and stoichiometric coefficients. Specifically, the sequence of reactions in the inorganic reaction information database should correspond to the reaction order based on the Gibbs free energy of the reactions\cite{gibbs}. The database includes 65 fundamental reactions\cite{keben}, categorized into four major types: Acid-Base, Double Displacement, Redox, and Complexation, as shown in Table 1.

The chemical substance information encompasses the symbol representation, color, enthalpy change, and state of the substances. The database includes 69 types of chemicals and ions, covering all substances involved in the aforementioned reactions, as listed in Table 2. Color is represented in RGB format, serving as the reference color for display along with the transparency value\cite{doi:10.1021/acs.jced.5b00018,wei2006}. Enthalpy change is given as the standard molar enthalpy of formation of the substances. States of substances include solid (s), liquid (l), gas (g), and solution (aq).

\renewcommand{\arraystretch}{1.2} 

\begin{table}[H]
    \centering
    \caption{Reactions}
    \label{tab:reaction}
    \begin{tabular}{ccc}
        \toprule
        Number & Reaction Type & Ionic Reaction Equation \\
        \midrule
        1 & Acid-Base & $\mathrm{H^+ + OH^- \rightarrow H_2O}$ \\
        2 & Acid-Base & $\mathrm{OH^- + HSO_4^- \rightarrow SO_4^{2-} + H_2O}$ \\
        3 & Acid-Base & $\mathrm{CH_3COOH + OH^- \rightarrow CH_3COO^- + H_2O}$ \\
        4 & Double Displacement & $\mathrm{2H^+ + Na_2O \rightarrow 2Na^+ + H_2O}$ \\
        5 & Double Displacement & $\mathrm{2H^+ + MgO \rightarrow Mg^{2+} + H_2O}$ \\
        6 & Double Displacement & $\mathrm{2H^+ + CaO \rightarrow Ca^{2+} + H_2O}$ \\
        7 & Double Displacement & $\mathrm{6H^+ + Al_2O_3 \rightarrow 2Al^{3+} + 3H_2O}$ \\
        8 & Double Displacement & $\mathrm{2H^+ + K_2O \rightarrow 2K^+ + H_2O}$ \\
        9 & Double Displacement & $\mathrm{2H^+ + FeO \rightarrow Fe^{2+} + H_2O}$ \\
        10 & Double Displacement & $\mathrm{2H^+ + CuO \rightarrow Cu^{2+} + H_2O}$ \\
        11 & Double Displacement & $\mathrm{2H^+ + ZnO \rightarrow Zn^{2+} + H_2O}$ \\
        12 & Double Displacement & $\mathrm{2H^+ + PbO \rightarrow Pb^{2+} + H_2O}$ \\
        13 & Double Displacement & $\mathrm{4H^+ + SnO_2 \rightarrow Sn^{4+} + 2H_2O}$ \\
        14 & Double Displacement & $\mathrm{2H^+ + Ag_2O \rightarrow 2Ag^+ + H_2O}$ \\
        15 & Double Displacement & $\mathrm{6H^+ + Fe_2O_3 \rightarrow 2Fe^{3+} + 3H_2O}$ \\
        16 & Redox & $\mathrm{5Fe^{2+} + MnO_4^- + 8H^+ \rightarrow 5Fe^{3+} + Mn^{2+} + 4H_2O}$ \\
        17 & Redox & $\mathrm{2MnO_4^- + 5H_2O_2 + 6H^+ \rightarrow 2Mn^{2+} + 5O_2 + 8H_2O}$ \\
        18 & Redox & $\mathrm{2MnO_4^- + 5SO_3^{2-} + H^+ \rightarrow 2Mn^{2+} + 5SO_4^{2-} + 3H_2O}$ \\
        19 & Redox & $\mathrm{2MnO_4^- + 5HSO_3^- + H^+ \rightarrow 2Mn^{2+} + 5SO_4^{2-} + 3H_2O}$ \\
        20 & Redox & $\mathrm{ClO^- + H_2O_2 \rightarrow Cl^- + O_2 + H_2O}$ \\
        \bottomrule
    \end{tabular}
\end{table}

\setcounter{table}{0}
\begin{table}[H]
    \centering
    \caption{Reactions}
    \label{tab:reaction}
    \begin{tabular}{ccc}
        \toprule
        Number & Reaction Type & Ionic Reaction Equation \\
        \midrule
        21 & Redox & $\mathrm{ClO^- + 2Fe^{2+} + 2H^+ \rightarrow Cl^- + 2Fe^{3+} + H_2O}$ \\
        22 & Redox & $\mathrm{ClO^- + SO_3^{2-} \rightarrow Cl^- + SO_4^{2-}}$ \\
        23 & Redox & $\mathrm{ClO^- + HSO_3^- \rightarrow Cl^- + SO_4^{2-} + H^+}$ \\
        24 & Redox & $\mathrm{2H_2O_2 + 2Fe^{2+} + 2H^+ \rightarrow 2Fe^{3+} + 2H_2O}$ \\
        25 & Redox & $\mathrm{SO_3^{2-} + H_2O_2 \rightarrow SO_4^{2-} + H_2O}$ \\
        26 & Redox & $\mathrm{HSO_3^- + H_2O_2 \rightarrow SO_4^{2-} + H^+ + H_2O}$ \\
        27 & Redox & $\mathrm{Mg + 2Ag^+ \rightarrow Mg^{2+} + 2Ag}$ \\
        28 & Redox & $\mathrm{Mg + Cu^{2+} \rightarrow Mg^{2+} + Cu}$ \\
        29 & Redox & $\mathrm{Mg + 2H^+ \rightarrow Mg^{2+} + H_2}$ \\
        30 & Redox & $\mathrm{Mg + Fe^{2+} \rightarrow Mg^{2+} + Fe}$ \\
        31 & Redox & $\mathrm{Mg + Zn^{2+} \rightarrow Mg^{2+} + Zn}$ \\
        32 & Redox & $\mathrm{3Mg + 2Al^{3+} \rightarrow 3Mg^{2+} + 2Al}$ \\
        33 & Redox & $\mathrm{Al + 3Ag^+ \rightarrow Al^{3+} + 3Ag}$ \\
        34 & Redox & $\mathrm{2Al + 3Cu^{2+} \rightarrow 2Al^{3+} + 3Cu}$ \\
        35 & Redox & $\mathrm{2Al + 6H^+ \rightarrow 2Al^{3+} + 3H_2}$ \\
        36 & Redox & $\mathrm{2Al + 3Zn^{2+} \rightarrow 2Al^{3+} + 3Zn}$ \\
        37 & Redox & $\mathrm{2Al + 3Fe^{2+} \rightarrow 2Al^{3+} + 3Fe}$ \\
        38 & Redox & $\mathrm{Zn + 2Ag^+ \rightarrow Zn^{2+} + 2Ag}$ \\
        39 & Redox & $\mathrm{Zn + Cu^{2+} \rightarrow Zn^{2+} + Cu}$ \\
        40 & Redox & $\mathrm{Zn + 2H^+ \rightarrow Zn^{2+} + H_2}$ \\
        41 & Redox & $\mathrm{Zn + Fe^{2+} \rightarrow Zn^{2+} + Fe}$ \\
        42 & Redox & $\mathrm{Fe + 2Ag^+ \rightarrow Fe^{2+} + 2Ag}$ \\
        43 & Redox & $\mathrm{Fe + 2Fe^{3+} \rightarrow 3Fe^{2+}}$ \\
        44 & Redox & $\mathrm{Fe + 2H^+ \rightarrow Fe^{2+} + H_2}$ \\
        45 & Redox & $\mathrm{Fe + Cu^{2+} \rightarrow Fe^{2+} + Cu}$ \\
        46 & Redox & $\mathrm{Cu + 2Ag^+ \rightarrow Cu^{2+} + 2Ag}$ \\
        47 & Double Displacement & $\mathrm{Ag^+ + Cl^- \rightarrow AgCl}$ \\
        48 & Double Displacement & $\mathrm{Ba^{2+} + SO_4^{2-} \rightarrow BaSO_4}$ \\
        49 &  Complexation & $\mathrm{Fe^{3+} + 3SCN^- \rightarrow Fe(SCN)_3}$ \\
        50 & Double Displacement & $\mathrm{Fe^{3+} + 3OH^- \rightarrow Fe(OH)_3}$ \\
        51 & Double Displacement & $\mathrm{Cu^{2+} + 2OH^- \rightarrow Cu(OH)_2}$ \\
        52 & Double Displacement & $\mathrm{Zn^{2+} + 2OH^- \rightarrow Zn(OH)_2}$ \\
        53 & Double Displacement & $\mathrm{Fe^{2+} + 2OH^- \rightarrow Fe(OH)_2}$ \\
        54 & Double Displacement & $\mathrm{Co^{2+} + 2OH^- \rightarrow Co(OH)_2}$ \\
        55 & Complexation & $\mathrm{Co^{2+} + 4Cl^- \rightarrow [CoCl_4]^{2-}}$ \\
        56 & Double Displacement & $\mathrm{Mn^{2+} + 2OH^- \rightarrow Mn(OH)_2}$ \\
        57 & Double Displacement & $\mathrm{Ca^{2+} + 2OH^- \rightarrow Ca(OH)_2}$ \\
        58 & Double Displacement & $\mathrm{Sn^{4+} + 4OH^- \rightarrow SnO_2 + 2H_2O}$ \\
        59 & Double Displacement & $\mathrm{2Ag^+ + 2OH^- \rightarrow Ag_2O + H_2O}$ \\
        60 & Double Displacement & $\mathrm{Al^{3+} + 3OH^- \rightarrow Al(OH)_3}$ \\
        61 & Double Displacement & $\mathrm{Al(OH)_3 + OH^- \rightarrow Al(OH)_4^-}$ \\
        62 & Complexation & $\mathrm{Pb^{2+} + 4Br^- \rightarrow [PbBr_4]^2-}$ \\
        63 & Double Displacement & $\mathrm{H^+ + SO_3^{2-} \rightarrow HSO_3^-}$ \\
        64 & Double Displacement & $ \mathrm{Pb^{2+} + SO_4^{2-} \rightarrow PbSO_4}$ \\
        65 & Double Displacement & $ \mathrm{Mg^{2+} + 2OH^- \rightarrow Mg(OH)_2} $ \\
        \bottomrule
    \end{tabular}
\end{table}

\renewcommand{\arraystretch}{1.25} 
\begin{table}[H]
    \centering
    \caption{Chemicals}
    \label{tab:chemicals}
    \begin{tabular}{c c c c c}        
    \toprule
				Number & Chemical & Color & Enthalpy & State \\
    \midrule
		1 & H\textsuperscript{+} & None & 0 & aq \\
		2 & OH\textsuperscript{-} & None & -229.9 & aq \\
		3 & H\textsubscript{2}O & None & -285.830 & l \\
		4 & Co\textsuperscript{2+} & \fcolorbox{black}[rgb]{0.81, 0.41, 0.55}{\texttt{(207, 104, 140)}} & -58.2 & aq \\
		5 & Co(OH)\textsubscript{2} & \fcolorbox{black}[rgb]{0.68, 0.85, 0.90}{\texttt{(173, 216, 230)}} & -539.7 & s \\
		6 & Fe\textsuperscript{3+} & \fcolorbox{black}[rgb]{1.00, 1.00, 0.60}{\texttt{(255, 255, 153)}} & -47.7 & aq \\
		7 & SCN\textsuperscript{-} & None & 76.44 & aq \\
		8 & Fe(SCN)\textsubscript{3} & \fcolorbox{black}[rgb]{0.59, 0.00, 0.09}{\textcolor{white}{\texttt{(150, 000, 024)}}} & None & s \\
		9 & Fe\textsuperscript{2+} & \fcolorbox{black}[rgb]{0.60, 0.70, 0.59}{\texttt{(152, 178, 150)}} & -87.9 & aq \\
		10 & SO\textsubscript{4}\textsuperscript{2-} & None & -907.5 & aq \\
		11 & K\textsuperscript{+} & None & -251.2 & aq \\
		12 & MnO\textsubscript{4}\textsuperscript{-} & \fcolorbox{black}[rgb]{0.50, 0.00, 0.50}{\textcolor{white}{\texttt{(128, 000, 128)}}} & -518.4 & aq \\
		13 & Mn\textsuperscript{2+} & \fcolorbox{black}[rgb]{0.65, 0.56, 0.75}{\texttt{(166, 142, 190)}} & -218.8 & aq \\
		14 & Cu\textsuperscript{2+} & \fcolorbox{black}[rgb]{0.00, 0.75, 1.00}{\texttt{(000, 191, 255)}} & 64.4 & s \\
		15 & Cu(OH)\textsubscript{2} & \fcolorbox{black}[rgb]{0.00, 0.00, 0.50}{\textcolor{white}{\texttt{(000, 000, 128)}}} & -450.37 & s \\
		16 & CoCl\textsubscript{4}\textsuperscript{2-} & \fcolorbox{black}[rgb]{1.00, 0.75, 0.80}{\texttt{(255, 192, 203)}} & None & s \\
		17 & Fe & \fcolorbox{black}[rgb]{0.75, 0.75, 0.75}{\texttt{(192, 192, 192)}} & 0 & s \\
		18 & Al & \fcolorbox{black}[rgb]{0.75, 0.75, 0.75}{\texttt{(192, 192, 192)}} & 0 & s \\
		19 & Al\textsuperscript{3+} & None & -524.7 & aq \\
		20 & AgCl & \fcolorbox{black}[rgb]{1.00, 1.00, 1.00}{\texttt{(255, 255, 255)}} & -127 & s \\
		21 & Ba\textsuperscript{2+} & None & -538.4 & aq \\
		22 & BaSO\textsubscript{4} & \fcolorbox{black}[rgb]{1.00, 1.00, 1.00}{\texttt{(255, 255, 255)}} & -1465.2 & s \\
		23 & Zn\textsuperscript{2+} & None & -152.4 & aq \\
		24 & Zn(OH)\textsubscript{2} & \fcolorbox{black}[rgb]{1.00, 1.00, 0.94}{\texttt{(255, 255, 240)}} & -642 & s \\
		25 & CH\textsubscript{3}COOH & None & -484.93 & aq \\
		26 & CH\textsubscript{3}COO\textsuperscript{-} & None & -496.4 & aq \\
		27 & Mn(OH)\textsubscript{2} & \fcolorbox{black}[rgb]{1.00, 1.00, 1.00}{\texttt{(255, 255, 255)}} & -700 & s \\
		28 & Fe(OH)\textsubscript{2} & \fcolorbox{black}[rgb]{0.00, 0.50, 0.00}{\texttt{(000, 128, 000)}} & -561.7 & s \\
		29 & Fe(OH)\textsubscript{3} & \fcolorbox{black}[rgb]{0.55, 0.27, 0.10}{\texttt{(139, 069, 019)}} & -824 & s \\
		30 & Ca\textsuperscript{2+} & None & -543.0 & aq \\
		31 & Ca(OH)\textsubscript{2} & \fcolorbox{black}[rgb]{1.00, 1.00, 1.00}{\texttt{(255, 255, 255)}} & -986.09 & s \\
		32 & Mg & \fcolorbox{black}[rgb]{0.75, 0.75, 0.75}{\texttt{(192, 192, 192)}} & 0 & s \\
		33 & Mg\textsuperscript{2+} & None & -462.0 & aq \\
		34 & Cu & \fcolorbox{black}[rgb]{0.55, 0.27, 0.10}{\texttt{(139, 069, 019)}} & 0 & s \\
		35 & Zn & \fcolorbox{black}[rgb]{0.75, 0.75, 0.75}{\texttt{(192, 192, 192)}} & 0 & s \\
		36 & Ag\textsuperscript{+} & None & 105.9 & s \\
		37 & Ag & \fcolorbox{black}[rgb]{1.00, 1.00, 1.00}{\texttt{(255, 255, 255)}} & 0 & s \\
		38 & H\textsubscript{2} & None & 0 & g \\
		39 & Pb\textsuperscript{2+} & None & 1.6 & aq \\
		40 & Br\textsuperscript{-} & None & -120.9 & aq \\
    		41 & [PbBr\textsubscript{4}]\textsuperscript{2-} & None & None & aq \\
		42 & I\textsuperscript{-} & None & -55.9 & aq \\
    \bottomrule
\end{tabular}
\end{table}

\setcounter{table}{1}
\begin{table}[H]
    \centering
    \caption{Chemicals}
    \label{tab:chemicals}
    \begin{tabular}{c c c c c}        
    \toprule
				Number & Chemical & Color & Enthalpy & State \\
    \midrule
    		43 & Na\textsubscript{2}O & \fcolorbox{black}[rgb]{1.00, 1.00, 1.00}{\texttt{(255, 255, 255)}} & -416 & s \\
		44 & Na\textsuperscript{+} & None & -239.7 & aq \\
		45 & MgO & \fcolorbox{black}[rgb]{1.00, 1.00, 1.00}{\texttt{(255, 255, 255)}} & -601.8 & s \\
    	46 & CaO & \fcolorbox{black}[rgb]{1.00, 1.00, 1.00}{\texttt{(255, 255, 255)}} & -635.5 & s \\
		47 & Al\textsubscript{2}O\textsubscript{3} & \fcolorbox{black}[rgb]{1.00, 1.00, 1.00}{\texttt{(255, 255, 255)}} & -1675.7 & s \\
    	48 & K\textsubscript{2}O & \fcolorbox{black}[rgb]{1.00, 1.00, 1.00}{\texttt{(255, 255, 255)}} & -363.17 & s \\
		49 & K\textsuperscript{+} & None & -251.2 & aq \\
		50 & FeO & \fcolorbox{black}[rgb]{0.00, 0.00, 0.00}{\textcolor{white}{\texttt{(000, 000, 000)}}} & -272.04 & s \\
		51 & CuO & \fcolorbox{black}[rgb]{0.00, 0.00, 0.00}{\textcolor{white}{\texttt{(000, 000, 000)}}} & -155.2 & s \\
		52 & ZnO & \fcolorbox{black}[rgb]{1.00, 1.00, 1.00}{\texttt{(255, 255, 255)}} & -348.0 & s \\
		53 & PbO & \fcolorbox{black}[rgb]{1.00, 1.00, 0.00}{\texttt{(255, 255, 000)}} & -217.9 & s \\
		54 & H\textsubscript{2}O\textsubscript{2} & None & -191.17 & aq \\
		55 & ClO\textsuperscript{-} & None & -107.6 & aq \\
		56 & Cl\textsuperscript{-} & None & -167.4 & aq \\
		57 & ClO\textsubscript{2}\textsuperscript{-} & None & -67.0 & aq \\
		58 & HSO\textsubscript{4}\textsuperscript{-} & None & -907.5 & aq \\
		59 & SO\textsubscript{3}\textsuperscript{2-} & None & -626.22 & aq \\
		60 & HSO\textsubscript{3}\textsuperscript{-} & None & -608.81 & aq \\
		61 & O\textsubscript{2} & None & 0 & g \\
		62 & Al(OH)\textsubscript{4}\textsuperscript{-} & None & -1502.9 & aq \\
		63 & SnO\textsubscript{2} & \fcolorbox{black}[rgb]{1.00, 1.00, 1.00}{\texttt{(255, 255, 255)}} & -580.7 & s \\
		64 & Sn\textsuperscript{4+} & None & 158.3 & aq \\
		65 & Al(OH)\textsubscript{3} & \fcolorbox{black}[rgb]{1.00, 1.00, 1.00}{\texttt{(255, 255, 255)}} & -1276 & s \\
		66 & Ag\textsubscript{2}O & \fcolorbox{black}[rgb]{0.55, 0.27, 0.10}{\texttt{(139, 069, 019)}} & -30.6 & s \\
		67 & Fe\textsubscript{2}O\textsubscript{3} & \fcolorbox{black}[rgb]{0.55, 0.27, 0.10}{\texttt{(139, 069, 019)}} & -822.2 & s \\
		68 & PbSO\textsubscript{4} & \fcolorbox{black}[rgb]{1.00, 1.00, 1.00}{\texttt{(255, 255, 255)}} & -911.36 & s \\
		69 & Mg(OH)\textsubscript{2} & \fcolorbox{black}[rgb]{1.00, 1.00, 1.00}{\texttt{(255, 255, 255)}} & -913.794 & s \\    \bottomrule
\end{tabular}
\end{table}

\renewcommand{\arraystretch}{1} 

\begin{figure}[H]
    \centering
    \includegraphics[width=0.8\textwidth]{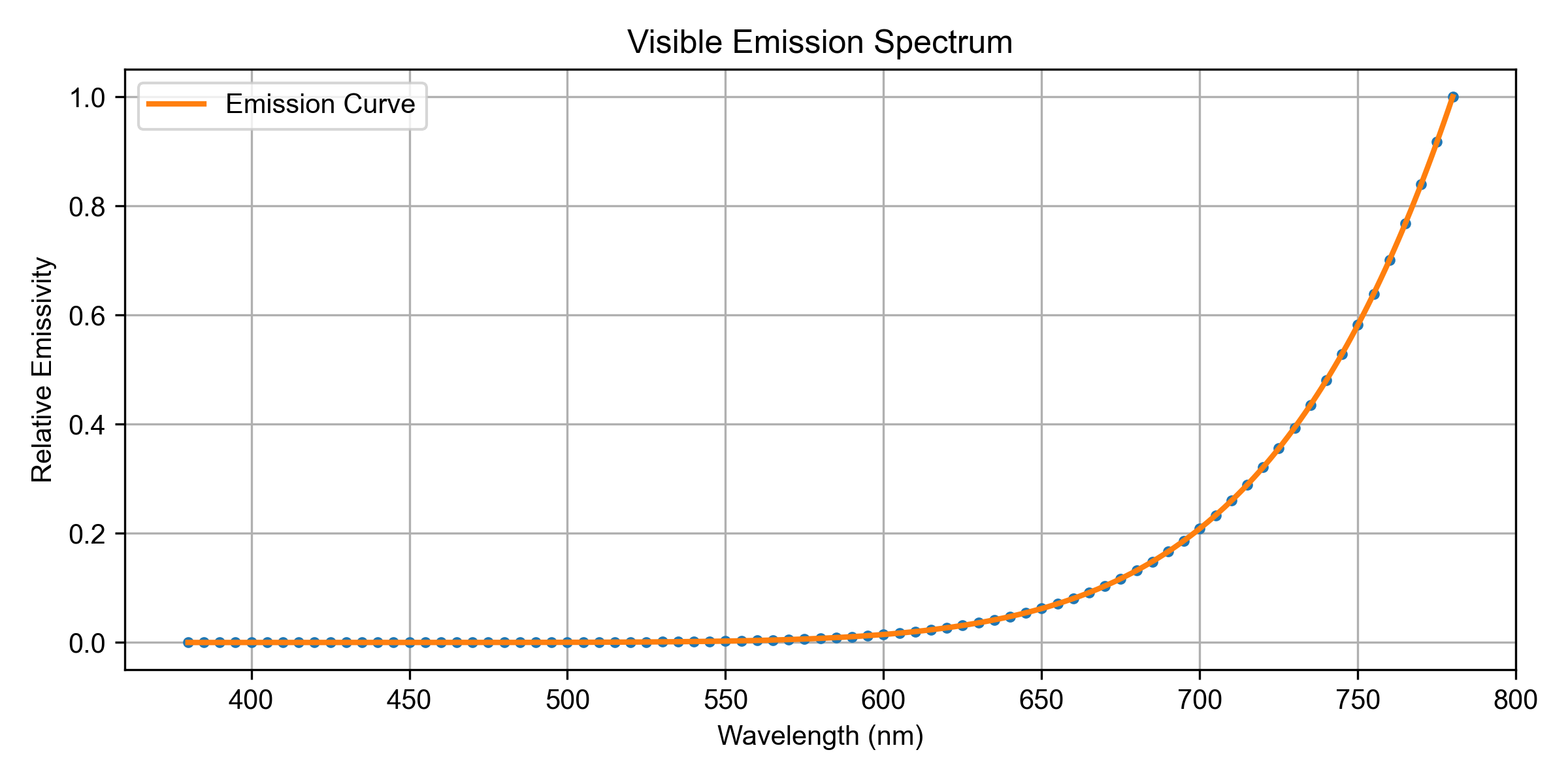}
    \caption{Visible emission spectrum of a blackbody radiator at 1000K}
    \label{fig:spectrum}
\end{figure}

\subsection{Spectrum to Color Conversion}
The simulator supports the conversion of visible spectra to RGB colors for substance characterization. The visible light spectrum of a substance determines its color appearance in a colorless transparent solution\cite{Udayakumar2014}. If the emission spectrum of a substance is provided, the simulator can calculate and convert the spectral data to the CIE color space and subsequently to the RGB color space. For example, the emission spectrum of a blackbody radiator at 1000K can be illustrated as Fig. \ref{fig:spectrum}. The yielding RGB color is calculates as \texttt{(255,2,0)}, which could be verified in other ways\cite{Fortner1997,cie}.

\subsection{Chemical Experiment}
\label{chemical_experiment}

\begin{figure}[H]
    \centering
    \includegraphics[width=\textwidth]{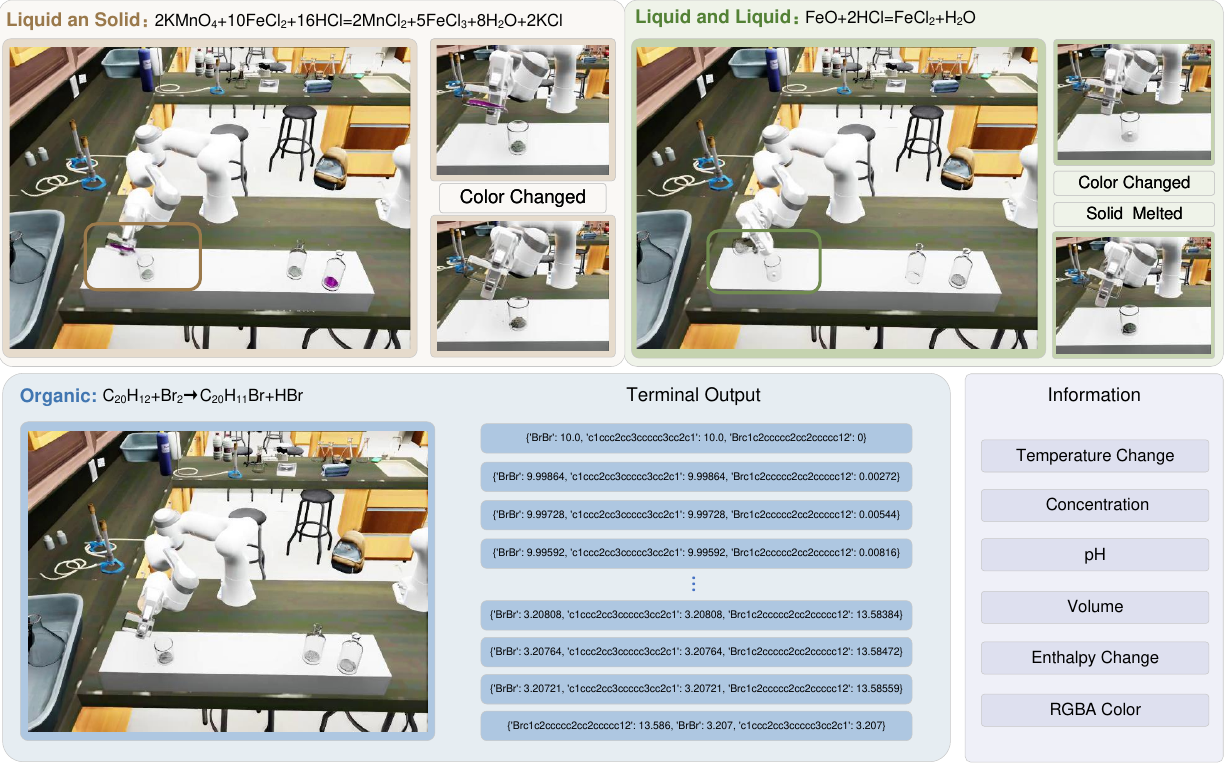}
    \caption{Illustration of the inorganic and organic reactions performed within the Chemistry3D environment. Inorganic reactions include the oxidation of KMnO\textsubscript{4} and FeCl\textsubscript{2}, and the dissolution of HCl and FeO. Users can observe significant color changes during the reaction between KMnO\textsubscript{4} and FeCl\textsubscript{2}, as well as during the dissolution and color change in the reaction between HCl and FeO. Organic reactions include those between compounds Br\textsubscript{2} and C\textsubscript{20}H\textsubscript{12}. Chemistry3D allows the output of intermediate states for both organic and inorganic reactions; however, detailed information on the reaction progress is available only for inorganic reactions.}
    \label{fig:reaction}
\end{figure}

In our chemical experiments, we focused on integrating robotic operations within the IsaacSim environment to enhance the visualization and analysis of both inorganic and organic reactions as shown in Fig.~\ref{fig:reaction}.

\textbf{Inorganic Experiment} The inorganic experiment selected the reactions of potassium permanganate (KMnO\textsubscript{4}) with ferrous chloride (FeCl\textsubscript{2}) and hydrochloric acid (HCl) with iron(II) oxide (FeO) as the subjects of investigation. In the simulation environment, a robotic arm was configured to perform tasks such as gripping and pouring. The robotic arm identified the location and labels of the objects within the scene and executed the corresponding chemical operations. During the inorganic experiment, Chemistry3D monitored the center of mass of the reactants to determine whether the reaction had occurred. It also provided the output of intermediate products after the reaction. Additionally, Chemistry3D supported the acquisition of data on the temperature, enthalpy change, and pH values of the reactants. In Chemistry3D, significant color changes were observed in the reaction between potassium permanganate and ferrous chloride, while the reaction between hydrochloric acid and iron(II) oxide exhibited the phenomenon of solid dissolution. These color changes and state information could be visualized in Chemistry3D. These two experiments demonstrated that Chemistry3D could realize and visualize reactions between solids and liquids as well as between liquids.



\textbf{Organic Experiment} In the organic experiment, most reactions do not exhibit significant color changes. The intermediate products in organic reactions are crucial for research. In Chemistry3D, we chose to simulate the formation of intermediate products in organic reactions. We adopted the same experimental setup as used in inorganic reactions to study the reaction between bromine (Br\textsubscript{2}) and pyrene (C\textsubscript{20}H\textsubscript{12}). During the simulation, the chemical simulation was synchronized with the robotic arm simulation, with detailed data on reactants and products at each step output to the terminal. Through this information, we can further optimize our reaction parameters and improve robotic manipulations to enhance the yield of the final product.


\begin{figure}[H]
    \centering
    \includegraphics[width=\textwidth]{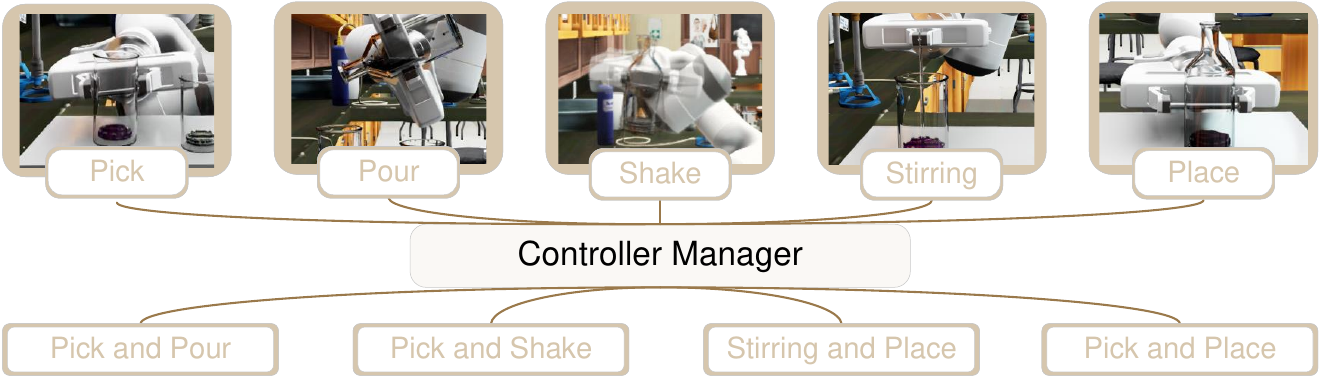}
    \caption{Illustration of motion types and bounding in Chemisrty3D.}
    \label{fig:Controller_Manager}
\end{figure}

\subsection{Chemical Operation}
\label{chemical_manipulation}

\begin{algorithm}
\caption{PlaceController}
\label{alg:PlaceController}
\begin{algorithmic}[1]
\STATE \textbf{class} PlaceController \textbf{inherits} BaseController
\STATE \textbf{def} \_\_init\_\_(name, cspace\_controller, gripper, end\_effector\_initial\_height, events\_dt, speed)
    \STATE Initialize parameters and phases
\STATE \textbf{def} forward(pour\_position, return\_position, current\_joint\_positions, end\_effector\_offset, end\_effector\_orientation)
    \IF{paused or is\_done()}
        \STATE pause
        \RETURN target\_joint\_positions with None values
    \ENDIF
    \IF{event == 3}
        \STATE Open the gripper
    \ELSE
        \IF{event == 0}
            \STATE Set current target x and y to pour position
        \ENDIF
        \STATE Interpolate xy between current and return positions
        \IF{event == 1}
            \STATE Move end effector to return position
        \ELSE
            \STATE Calculate target height and move end effector
        \ENDIF
    \ENDIF
    \STATE Update event and time
    \RETURN target\_joint\_positions

\end{algorithmic}
\end{algorithm}

Chemical manipulation typically involves specified operations on chemical containers, such as gripping, pouring, stirring, shaking, and placing. In Chemistry3D, we have modularly designed these chemical operations to enable the completion of specific, comprehensive chemical tasks. As shown in Fig.~\ref{fig:Controller_Manager}, We developed a class called Controller Manager to manage these operations. In our setup, each chemical operation is associated with a corresponding controller (Controller). For each instantiated Controller, the Controller Manager ensures that individual chemical operations are performed sequentially and in conjunction. In Sections \ref{chemical_experiment}, we have integrated the tasks of gripping, pouring, and placing through the Controller Manager, achieving a modular design for robotic manipulations.

To illustrate, consider the process of placing an object, which can be decomposed into five distinct stages: vertical ascent, horizontal translation, vertical descent, gripper release, and a subsequent vertical ascent. This sequence assumes that the robotic arm maintains a secure grip on the object throughout the entire placing action. As outlined in Algorithm~\ref{alg:PlaceController}, the entire motion process is segmented into multiple phases, with distinct time steps assigned to each phase to achieve adjustable motion speeds at different stages. Notably, to ensure stable operation of the robotic arm during the placing or grasping processes, intermediate points are incorporated within the vertical ascent and descent phases. This approach is designed to prevent deviation from the predetermined trajectory, thereby minimizing the risk of collision.

After defining a specific controller, we introduce the \textit{Controller Manager} class, which is responsible for managing all the controllers. The \texttt{add\_controller()} method is used to add instantiated controllers, while the \texttt{add\_task()} method specifies the execution order of tasks. Finally, the \texttt{execute()} method is employed to run all the controllers, thereby driving the robotic arm.

\begin{algorithm}
\caption{add\_controller Method}
\label{alg:add_controller}
\begin{algorithmic}[1]
\STATE \textbf{class} ControllerManager:
\STATE \textbf{def} add\_controller(controller\_name, controller\_instance):
    \IF{controller\_name already exists}
        \STATE add index to make the controller\_name unique
    \ENDIF
    \STATE append controller\_name and controller\_instance to the controllers list
\end{algorithmic}
\end{algorithm}

First, the \texttt{add\_controller} method is defined in Algorithm~\ref{alg:add_controller}. This method is responsible for adding a new controller instance to the list of controllers managed by the \texttt{ControllerManager}. The method takes two arguments: \texttt{controller\_name}, which is a string representing the name of the controller, and \texttt{controller\_instance}, which is the instance of the controller to be added.

The method first checks if a controller with the same name already exists in the list of controllers. If it does, it adds an index to the name to ensure uniqueness. Once a unique name is ensured, the method appends a dictionary containing the \texttt{controller\_name} and \texttt{controller\_instance} to the \texttt{controllers} list.

\begin{algorithm}
\caption{add\_task Method}
\label{alg:add_task}
\begin{algorithmic}[1]
\STATE \textbf{def} add\_task(controller\_type, param\_template):
    \STATE append controller\_type and param\_template to the tasks list
\end{algorithmic}
\end{algorithm}

Next, we have the \texttt{add\_task} method, which is shown in Algorithm~\ref{alg:add_task}. This method is used to add a new task to the list of tasks managed by the \texttt{ControllerManager}. The method takes two arguments: \texttt{controller\_type}, which is a string representing the type of the controller for the task, and \texttt{param\_template}, which is a dictionary containing the template for the task parameters. This ensures that the robotic arm performs the tasks in the correct order.

\begin{algorithm}
\caption{execute Method}
\label{alg:execute}
\begin{algorithmic}[1]
\STATE \textbf{def} execute(current\_observations):
    \IF{all tasks are completed}
        \STATE print "All tasks completed"
        \STATE pause the world simulation
        \STATE \textbf{return}
    \ENDIF
    
    \STATE task = get current task from the tasks list
    \STATE controller = get current controller from the controllers list
    
    \STATE ensure the controller exists:
    \IF{not}
        \STATE print "Controller not found!"
        \STATE \textbf{return}
    \ENDIF
    
    \STATE generate task parameters from param\_template using current\_observations
    
    \STATE actions = call controller's forward method with generated task parameters
    
    \IF{controller\_type is 'pour'}
        \IF{controller's reaction started}
            \STATE update reactant information based on controller status
        \ENDIF
    \ENDIF

    \STATE apply actions to the franka robot

    \IF{controller is done}
        \STATE move to the next task by incrementing current\_task\_index
    \ENDIF
\end{algorithmic}
\end{algorithm}

Finally, the \texttt{execute} method is detailed in Algorithm~\ref{alg:execute}. This method is responsible for executing the current task using the current controller. It takes one argument: \texttt{current\_observations}, which is a dictionary containing the current observations from the simulation.

The method first checks if all tasks are completed. If they are, it prints a message and pauses the simulation. If not, it retrieves the current task and controller. It then ensures that the controller exists. If the controller does not exist, it prints an error message and returns.

Next, the method generates task parameters using the \texttt{current\_observations}. It then calls the \texttt{forward} method of the controller with the generated task parameters to get the actions.

If the controller type is 'pour', the method performs additional checks and updates related to the pouring process, such as checking if the reaction has been activated, and updating the simulation container color.

Finally, the method applies the actions to the robot and checks if the controller is done. If the controller is done, it moves to the next task by incrementing the \texttt{current\_task\_index}

Utilizing the Controller and ControllerManager frameworks, we have achieved a modular approach to managing robotic arm movements and control operations. This design enhances the maintainability and extensibility of the codebase, facilitating efficient and flexible transitions between different controllers and task allocations.


\subsection{Visual Sim2Real}
\label{Transparency}
\begin{figure}[H]
    \centering
    \includegraphics[width=\textwidth]{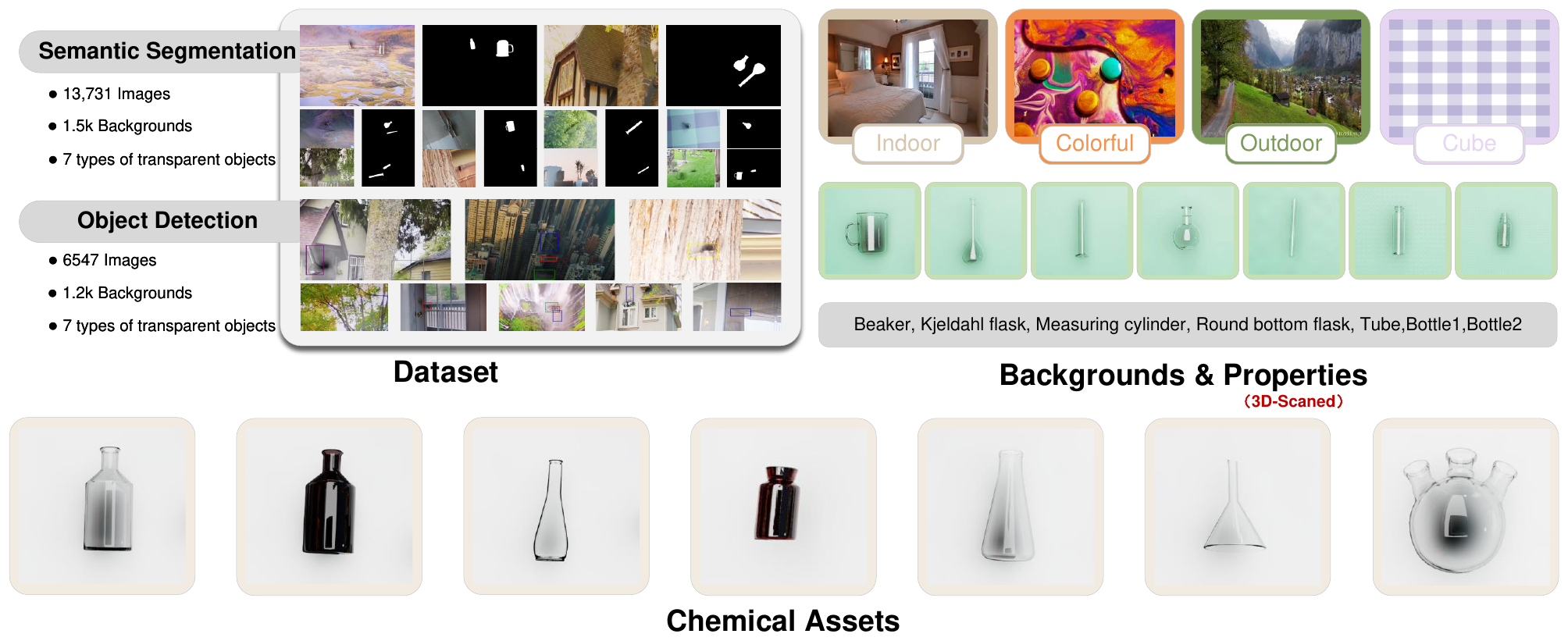}
    \caption{Dataset and chemical assets of Chemistry3D.}
    \label{fig:Dataset}
\end{figure}

Within the Omniverse platform, we selected transparent objects commonly found in chemical laboratories and introduced 3D-scanned transparent models. To verify the efficacy of transparent object simulation for direct Sim2Real transfer, we constructed two simulation datasets of transparent objects as shown in Fig.~\ref{fig:Dataset} focused on Semantic Segmentation and Target Detection tasks.

\begin{figure}[H]
    \centering
    \includegraphics[width=\textwidth]{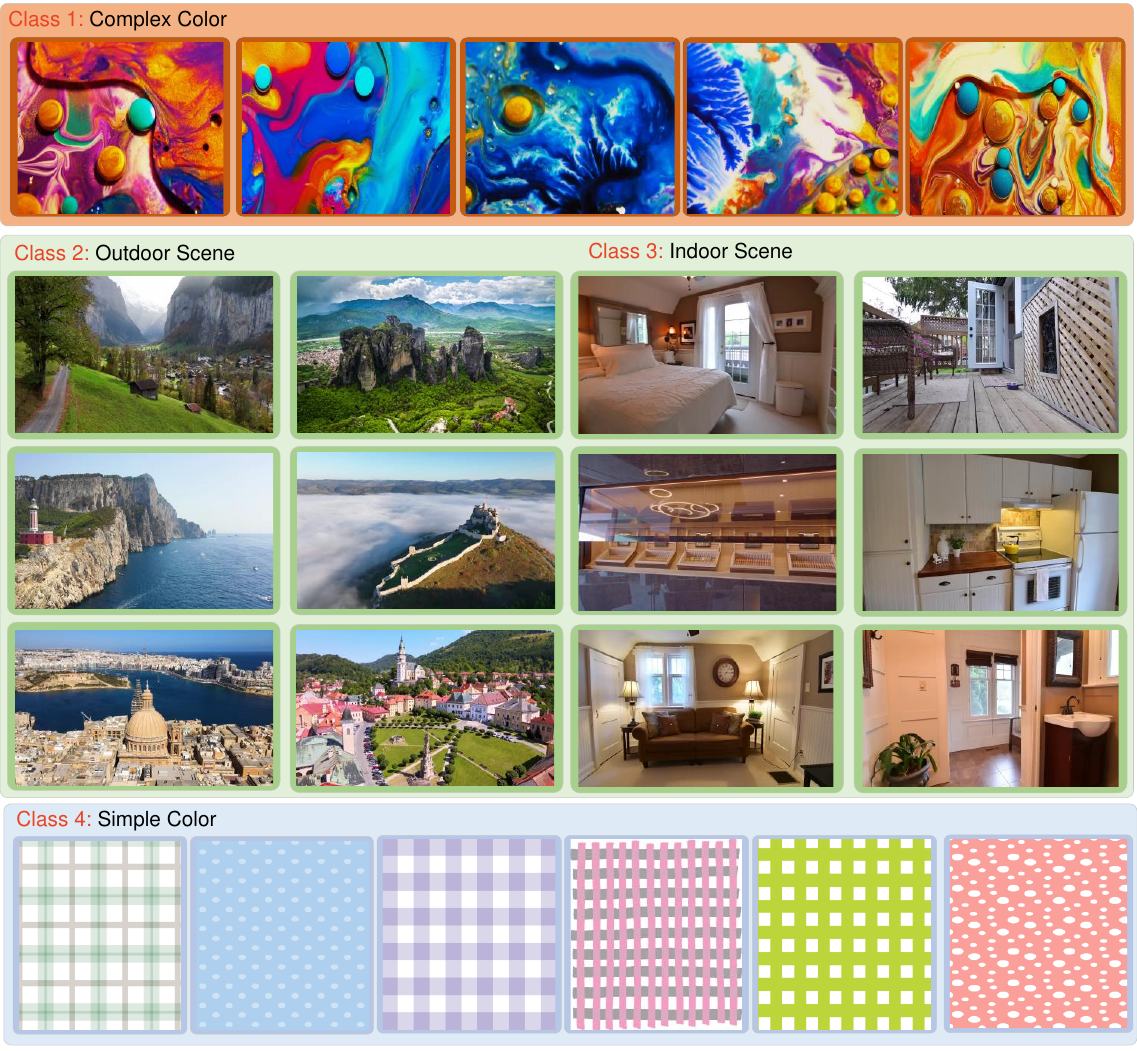}
    \caption{Illustration of backgrounds used in Dataset. These backgrounds includes complex colors, outdoor scenes, indoor scenes, and simple colors, were used to improve the generalization of the model for transparent object detection.}
    \label{fig:Backgrounds}
\end{figure}

In our dataset production process, we utilized seven real transparent objects scanned in 3D and applied glass material using Omniverse. As illustrated in the Fig.~\ref{fig:Backgrounds} , we incorporated a variety of complex colors, outdoor scenes, indoor scenes, and simple colors, resulting in a total of over 1,500 backgrounds. This approach was designed to ensure a sufficiently complex and diverse set of backgrounds. To enhance the dataset's generalizability, we dynamically adjusted the number of light sources, light intensity, and color temperature, aiming to maximize the number of reflective spots on the transparent objects during illumination. Specifically, we used three light sources to illuminate the objects, intentionally creating as many reflective spots as possible on their surfaces. Additionally, we applied randomized settings to the position and angle of the transparent objects, placing single or multiple objects in each image to enhance complexity. The Fig.~\ref{fig:Dataset} presents a sample of the generated dataset.

\begin{figure}
    \centering
    \includegraphics[width=\textwidth]{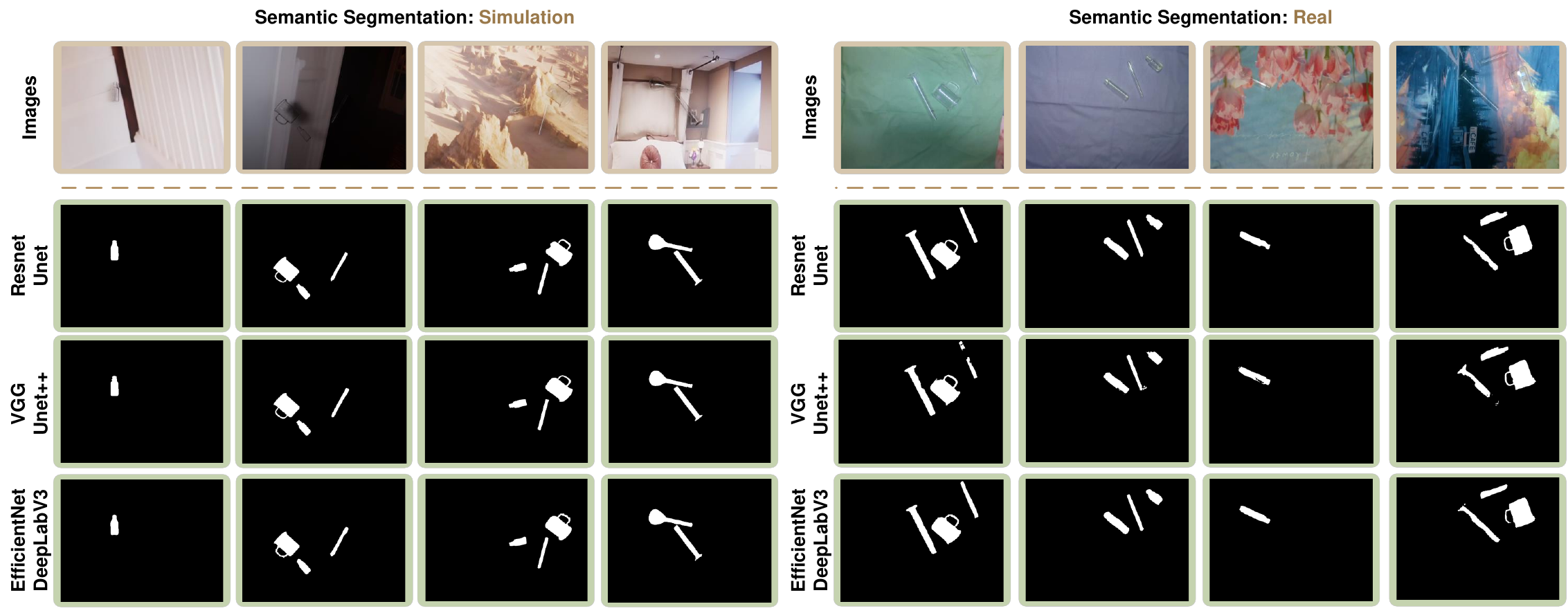}
    \caption{Illustration of the Sim2Real effect on the semantic segmentation task.}
    \label{fig:bench}
\end{figure}

The performance of Sim2Real is highly dependent on the quality of the simulated environment's visualization. Utilizing the high-fidelity visualization capabilities provided by Omniverse, we aim to demonstrate the Sim2Real capabilities for transparent objects achievable with Chemistry3D. We trained several mainstream encoder-decoder networks within the simulation environment and deployed them directly into real-world scenes to compare their semantic segmentation performance with that of TGCNN\cite{tgcnn}. Our network combinations include ResNet\cite{resnet} with UNet\cite{unet}, VGG\cite{VGG} with UNet++\cite{unet++}, and EfficientNet\cite{efficientnet} with DeepLabV3\cite{DeepLabV3}, each known for their robust performance in various segmentation tasks. As illustrated in Fig.~\ref{fig:bench}, these networks maintain robust segmentation capabilities across both simple and complex backgrounds in real-world environments. Chemistry3D also supports quantitative comparisons of network performance.

\begin{figure}[H]
    \centering
    \includegraphics[width=\textwidth]{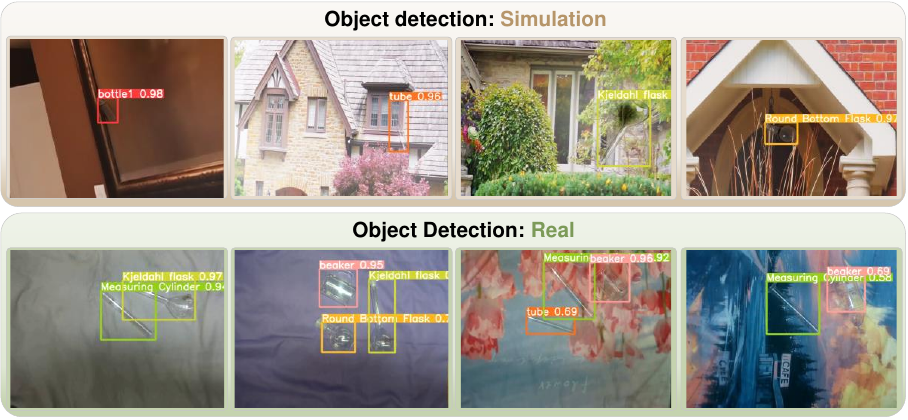}
    \caption{Illustration of the Sim2Real effect on the object detection task. }
    \label{fig:Object_detection}
\end{figure}

Additionally, we further validated the Sim2Real ability of vision tasks oriented towards transparent objects by performing a object detection task. For this purpose, we selected the YOLO\cite{yolo} algorithm. The performance of the network, trained within the simulation environment, was evaluated for target detection in both simulated and real-world environments. As illustrated in Fig.~\ref{fig:Object_detection}, the results are consistent with those observed in the semantic segmentation task, confirming that Chemistry3D effectively supports Sim2Real for transparent-object vision tasks.

\begin{figure}[H]
    \centering
    \includegraphics[width=\textwidth]{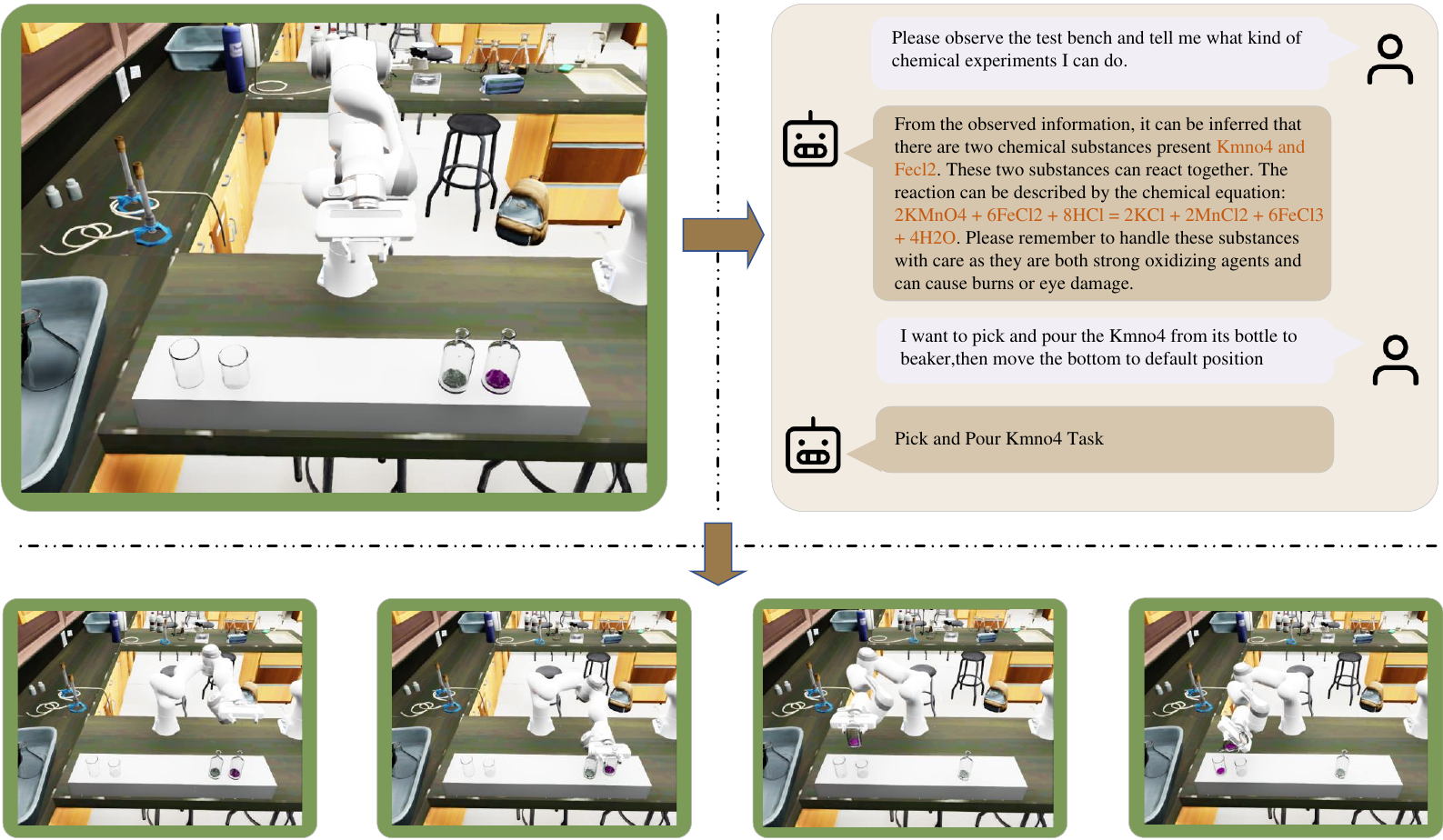}
    \caption{Embodied intelligence work in Chemistry3D. Each specific agent responds to the natural language, driving the movement of the robotic arm}
    \label{fig:Embodied}
\end{figure}

\subsection{Embodied Intelligence}
\label{Embodied}
To evaluate the development capabilities of Chemistry3D in embodied intelligence tasks, we initially designed a chemical experiment scene. The laboratory setup included a table equipped with containers of KMnO\textsubscript{4} and FeCl\textsubscript{2}, as well as two empty beakers. Within the overall framework, we constructed agents for robotic control. These agents were responsible for acquiring environmental information, generating robotic operation tasks, initializing different motion controllers, and managing robotic operations through the Controller Manager. The agents acquired information about the experimental scene, enabling the robot to observe interactive objects and generate potential chemical reactions based on its chemical reaction knowledge base. Subsequently, as shown in Fig.~\ref{fig:Embodied}, we utilized natural language input to direct the robot to complete the relevant chemical experiment tasks.

\begin{figure}[H]
    \centering
    \includegraphics[width=\textwidth]{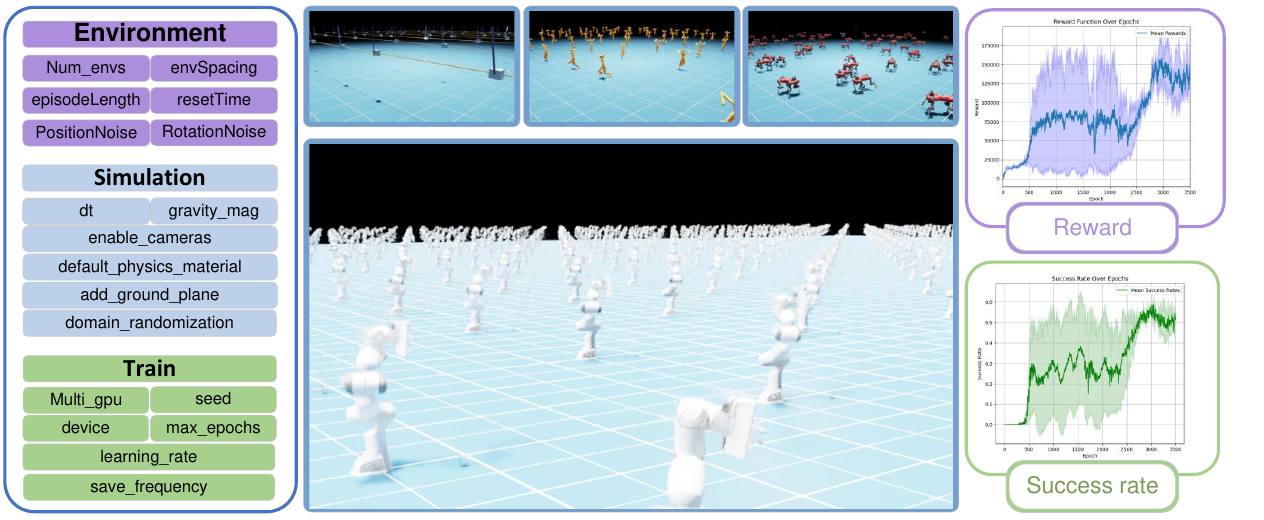}
    \caption{Reinforcement learning tasks in Chemistry3D. Reinforcement learning tasks support the tuning of hyperparameters for Environment, Simulation and Train.}
    \label{fig:RL}
\end{figure}

\subsection{Reinforcement Learning}
\label{Reinforcement}
OmniIsaacGymEnvs, integrated within IsaacSim, facilitates complex reinforcement learning tasks in Chemistry3D. To demonstrate the potential for developing reinforcement learning tasks in Chemistry3D in our experiment, we designated picking as the RL task as shown in Fig.~\ref{fig:RL}. We utilized a reward function setup similar to the provided examples, and employed Proximal Policy Optimization (PPO)\cite{PPO} as the reinforcement learning algorithm. In the reinforcement learning experiment, we configured the number of environments to 2048, the number of epochs to 3500, and the learning rate to  \(5 \times 10^{-4}\). In OmniIsaacGymEnvs, various hyperparameters(HP) related to simulation, environment, and training can be adjusted to optimize the overall effect of reinforcement learning training. Table ~\ref{tab:simulation_parameters} presents a selection of these HP along with the default configurations employed in our task. This customization allows for a more tailored approach to reinforcement learning, potentially enhancing the training outcomes. This training configuration was consistently applied across three separate realizations. As illustrated in the Fig.~\ref{fig:RL}, we plotted the reward curve and the success rate curve during training, using the average of the three realizations as the central curve and the standard deviation among the three experiments to represent the curve width. The robustness of the reinforcement learning task outcomes is evident from these results. Fig.~\ref{fig:RL} illustrates the training process, including the reward curve and success rate information for the grasping task.

\begin{longtable}{ll ll ll}
\caption{Simulation Environment and Training Parameters}
\label{tab:simulation_parameters} \\
\toprule
\textbf{Category} & \textbf{HP} & 
\textbf{Category} & \textbf{HP} & \textbf{Category} & \textbf{HP} \\
\midrule
\textbf{Environment} &  & \textbf{Simulation} &  &
\textbf{Train} & \\
Num\_envs & 2048  & domain\_randomization &  False & Multi\_gpu & True\\
envSpacing & 3.0  & dt & 1/120.0 
& seed & 42 \\
episodeLength & 500 & default\_physics\_material: &  & device & gpu \\
RotationNoise & 0.0 & \quad static\_friction & 1.0 & learning\_rate & $5 \times 10^{-4}$ \\
PositionNoise & 0.0 & \quad dynamic\_friction & 1.0 & max\_epochs & 3500\\
 &  & \quad restitution & 0.0 & save\_frequency & 100 \\
 &  & add\_ground\_plane & True \\
 &  & gravity\_mag & -9.81 \\
\bottomrule
\end{longtable}

{
\small
\bibliographystyle{unsrt}
\bibliography{ref}

\begin{thebibliography}{10}

\bibitem{leardi2009experimental}
Riccardo Leardi.
\newblock Experimental design in chemistry: A tutorial.
\newblock {\em Analytica Chimica Acta}, 652(1-2):161--172, 2009.

\bibitem{engineer}
The average work hours for a chemical engineer.
\newblock https://www.comsol.com/.

\bibitem{consumption}
Global consumption of chemicals in 2022.
\newblock https://www.statista.com/statistics/486582/worldwide-consumption-of-chemicals-in-by-region/.

\bibitem{van1990computer}
Wilfred~F Van~Gunsteren and Herman~JC Berendsen.
\newblock Computer simulation of molecular dynamics: methodology, applications, and perspectives in chemistry.
\newblock {\em Angewandte Chemie International Edition in English}, 29(9):992--1023, 1990.

\bibitem{dimian2014integrated}
Alexandre~C Dimian, Costin~Sorin Bildea, and Anton~A Kiss.
\newblock {\em Integrated design and simulation of chemical processes}.
\newblock Elsevier, 2014.

\bibitem{sun2018pyscf}
Qiming Sun, Timothy~C Berkelbach, Nick~S Blunt, George~H Booth, Sheng Guo, Zhendong Li, Junzi Liu, James~D McClain, Elvira~R Sayfutyarova, Sandeep Sharma, et~al.
\newblock Pyscf: the python-based simulations of chemistry framework.
\newblock {\em Wiley Interdisciplinary Reviews: Computational Molecular Science}, 8(1):e1340, 2018.

\bibitem{motard1975steady}
RL~Motard, M~Shacham, and EM~Rosen.
\newblock Steady state chemical process simulation.
\newblock {\em AIChE Journal}, 21(3):417--436, 1975.

\bibitem{yuan2022pre}
Zhecheng Yuan, Zhengrong Xue, Bo~Yuan, Xueqian Wang, Yi~Wu, Yang Gao, and Huazhe Xu.
\newblock Pre-trained image encoder for generalizable visual reinforcement learning.
\newblock {\em Advances in Neural Information Processing Systems}, 35:13022--13037, 2022.

\bibitem{granda2018controlling}
Jaros{\l}aw~M Granda, Liva Donina, Vincenza Dragone, De-Liang Long, and Leroy Cronin.
\newblock Controlling an organic synthesis robot with machine learning to search for new reactivity.
\newblock {\em Nature}, 559(7714):377--381, 2018.

\bibitem{de2019synthetic}
A~Filipa de~Almeida, Rui Moreira, and Tiago Rodrigues.
\newblock Synthetic organic chemistry driven by artificial intelligence.
\newblock {\em Nature Reviews Chemistry}, 3(10):589--604, 2019.

\bibitem{lodewyk2012computational}
Michael~W Lodewyk, Matthew~R Siebert, and Dean~J Tantillo.
\newblock Computational prediction of 1h and 13c chemical shifts: a useful tool for natural product, mechanistic, and synthetic organic chemistry.
\newblock {\em Chemical Reviews}, 112(3):1839--1862, 2012.

\bibitem{beeler2023chemgymrl}
Chris Beeler, Sriram~Ganapathi Subramanian, Kyle Sprague, Colin Bellinger, Mark Crowley, and Isaac Tamblyn.
\newblock Chemgymrl: An interactive framework for reinforcement learning for digital chemistry.
\newblock In {\em NeurIPS 2023 AI for Science Workshop}, 2023.

\bibitem{rajak2021autonomous}
Pankaj Rajak, Aravind Krishnamoorthy, Ankit Mishra, Rajiv Kalia, Aiichiro Nakano, and Priya Vashishta.
\newblock Autonomous reinforcement learning agent for chemical vapor deposition synthesis of quantum materials.
\newblock {\em npj Computational Materials}, 7(1):108, 2021.

\bibitem{zhang2021deep}
Jun Zhang, Yao-Kun Lei, Zhen Zhang, Xu~Han, Maodong Li, Lijiang Yang, Yi~Isaac Yang, and Yi~Qin Gao.
\newblock Deep reinforcement learning of transition states.
\newblock {\em Physical Chemistry Chemical Physics}, 23(11):6888--6895, 2021.

\bibitem{sridharan2024deep}
Bhuvanesh Sridharan, Animesh Sinha, Jai Bardhan, Rohit Modee, Masahiro Ehara, and U~Deva Priyakumar.
\newblock Deep reinforcement learning in chemistry: A review.
\newblock {\em Journal of Computational Chemistry}, 2024.

\bibitem{zhou2017optimizing}
Zhenpeng Zhou, Xiaocheng Li, and Richard~N Zare.
\newblock Optimizing chemical reactions with deep reinforcement learning.
\newblock {\em ACS Central Science}, 3(12):1337--1344, 2017.

\bibitem{jiang2023robotic}
Jiaqi Jiang, Guanqun Cao, Jiankang Deng, Thanh-Toan Do, and Shan Luo.
\newblock Robotic perception of transparent objects: A review.
\newblock {\em IEEE Transactions on Artificial Intelligence}, 2023.

\bibitem{gasteiger1990models}
Johann Gasteiger, Mario Marsili, MG~Hutchings, Heinz Saller, Peter Loew, P~R{\"o}se, and K~Rafeiner.
\newblock Models for the representation of knowledge about chemical reactions.
\newblock {\em Journal of Chemical Information and Computer Sciences}, 30(4):467--476, 1990.

\bibitem{engkvist2018computational}
Ola Engkvist, Per-Ola Norrby, Nidhal Selmi, Yu-hong Lam, Zhengwei Peng, Edward~C Sherer, Willi Amberg, Thomas Erhard, and Lynette~A Smyth.
\newblock Computational prediction of chemical reactions: current status and outlook.
\newblock {\em Drug Discovery Today}, 23(6):1203--1218, 2018.

\bibitem{omniverse}
{Omniverse}.
\newblock https://www.nvidia.com/en-us/omniverse/.

\bibitem{hummel2019leveraging}
Mathias Hummel and Kees van Kooten.
\newblock Leveraging nvidia omniverse for in situ visualization.
\newblock In {\em High Performance Computing: ISC High Performance 2019 International Workshops, Frankfurt, Germany, June 16-20, 2019, Revised Selected Papers 34}, pages 634--642. Springer, 2019.

\bibitem{ic}
{InteractiveChemistry}: Games and simulations.
\newblock https://interactivechemistry.org.

\bibitem{cfd}
Wenbin Li, K.~Yu, B.~Liu, and X.~Yuan.
\newblock Computational fluid dynamics simulation of hydrodynamics and chemical reaction in a {CFB} downer.
\newblock {\em Powder Technology}, 269:425--436, 01 2015.

\bibitem{molecule}
Zhenqin Wu, Bharath Ramsundar, Evan N. Feinberg, Joseph Gomes, Caleb Geniesse, Aneesh~S. Pappu, Karl Leswing, and Vijay Pande.
\newblock {MoleculeNet}: a benchmark for molecular machine learning.
\newblock {\em Chemical Science}, 9:513--530, 2018.

\bibitem{ord}
Steven Kearnes, Michael Maser, Michael Wleklinski, Anton Kast, Abigail Doyle, Spencer Dreher, Joel Hawkins, Klavs Jensen, and Connor Coley.
\newblock The open reaction database.
\newblock {\em Journal of the American Chemical Society}, 11 2021.

\bibitem{orderly}
Daniel~S Wigh, Joe Arrowsmith, Alexander Pomberger, Kobi~C Felton, and Alexei~A Lapkin.
\newblock {ORDerly}: Data sets and benchmarks for chemical reaction data.
\newblock {\em Journal of Chemical Information and Modeling}, 64(9):3790—3798, May 2024.

\bibitem{chemspider}
Harry Pence and Antony Williams.
\newblock Chemspider: An online chemical information resource.
\newblock {\em Journal of Chemical Education}, 87, 08 2010.

\bibitem{chemaxon}
Cheminformatics software for the next generation of scientists.
\newblock https://chemaxon.com.

\bibitem{rxn}
Rxn for chemistry.
\newblock https://rxn.res.ibm.com/rxn/.

\bibitem{nature}
Philippe Schwaller, Daniel Probst, Alain~C. Vaucher, Vishnu~H. Nair, David Kreutter, Teodoro Laino, and Jean‐Louis Reymond.
\newblock Mapping the space of chemical reactions using attention-based neural networks.
\newblock {\em Nature Machine Intelligence}, 3:144 -- 152, 2020.

\bibitem{chemreax}
Chemreax: a chemical reaction modeling and simulation app from sciencebysimulation.
\newblock https://www.sciencebysimulation.com/chemreax/Analyzer.aspx.

\bibitem{rah}
Caleb Chuck, Carl Qi, Michael~J. Munje, Shuozhe Li, Max Rudolph, Chang Shi, Siddhant Agarwal, Harshit Sikchi, Abhinav Peri, Sarthak Dayal, Evan Kuo, Kavan Mehta, Anthony Wang, Peter Stone, Amy Zhang, and Scott Niekum.
\newblock Robot air hockey: A manipulation testbed for robot learning with reinforcement learning, 2024.

\bibitem{mujoco}
Zichun Xu, Yuntao Li, Xiaohang Yang, Zhiyuan Zhao, Lei Zhuang, and Jingdong Zhao.
\newblock Open-source reinforcement learning environments implemented in mujoco with franka manipulator, 2024.

\bibitem{unity}
Shu-Guang Ouyang, Gang Wang, Jun-Yan Yao, Guang-Heng-Wei Zhu, Zhao-Yue Liu, and Chi Feng.
\newblock A {Unity3D}-based interactive three-dimensional virtual practice platform for chemical engineering.
\newblock {\em Computer Applications in Engineering Education}, 26, 08 2017.

\bibitem{cabd}
Le~Zou, Ze-Sheng Ding, Shuo-Yi Ran, Zhi-Ze Wu, Yun-Sheng Wei, Zhi-Huang He, and Xiao-Feng Wang.
\newblock A benchmark dataset in chemical apparatus: recognition and detection.
\newblock {\em Multimedia Tools and Applications}, 83:1--19, 08 2023.

\bibitem{BENNETT2022100831}
Jeffrey~A Bennett and Milad Abolhasani.
\newblock Autonomous chemical science and engineering enabled by self-driving laboratories.
\newblock {\em Current Opinion in Chemical Engineering}, 36:100831, 2022.

\bibitem{chemos}
Loïc~M. Roch, Florian Häse, and Alán Aspuru-Guzik.
\newblock {ChemOS}: An orchestration software to democratize autonomous discovery.
\newblock In {\em {Artificial Intelligence in Drug Discovery}}. The Royal Society of Chemistry, 11 2020.

\bibitem{archemist}
Hatem Fakhruldeen, Gabriella Pizzuto, Jakub Glowacki, and Andrew~Ian Cooper.
\newblock {ARChemist}: Autonomous robotic chemistry system architecture.
\newblock In {\em 2022 International Conference on Robotics and Automation (ICRA)}, pages 6013--6019, 2022.

\bibitem{Giraud1978}
Alain Giraud and Michel Petit.
\newblock {\em Chemistry of Charge Conservation}, pages 136--174.
\newblock Springer Netherlands, Dordrecht, 1978.

\bibitem{keben}
{Theodore L.} Brown, {H. Eugene} LeMay, {Bruce Edward} Bursten, {Catherine J.} Murphy, and {Patrick M.} Woodward.
\newblock {\em Chemistry: The Central Science}.
\newblock Pearson Prentice Hall, 12 edition, 2012.

\bibitem{Heald1974}
C.~Heald and A.~C.~K. Smith.
\newblock {\em Ionic Reactions and Electrochemical Methods of Analysis}, pages 278--333.
\newblock Macmillan Education UK, London, 1974.

\bibitem{doi:10.1021/j100324a007}
Bruce~J. Berne, Michal Borkovec, and John~E. Straub.
\newblock Classical and modern methods in reaction rate theory.
\newblock {\em The Journal of Physical Chemistry}, 92(13):3711--3725, 1988.

\bibitem{doi:10.1021/acs.jced.5b00018}
Chan-Yuan Tan and Yao-Xiong Huang.
\newblock Dependence of refractive index on concentration and temperature in electrolyte solution, polar solution, nonpolar solution, and protein solution.
\newblock {\em Journal of Chemical \& Engineering Data}, 60(10):2827--2833, 2015.

\bibitem{wei2006}
Wen~Li Wei and Xiu~Fang Yang.
\newblock Research on liquid concentration real-time detecting system based on f-p interferometer.
\newblock In {\em Experimental Mechanics in Nano and Biotechnology}, volume 326 of {\em Key Engineering Materials}, pages 143--146. Trans Tech Publications Ltd, 12 2006.

\bibitem{Sundararajan2017}
D.~Sundararajan.
\newblock {\em Color Image Processing}, pages 407--438.
\newblock Springer Singapore, Singapore, 2017.

\bibitem{Udayakumar2014}
Neetha Udayakumar.
\newblock {\em Visible Light Imaging}, pages 67--86.
\newblock Springer Berlin Heidelberg, Berlin, Heidelberg, 2014.

\bibitem{Fortner1997}
Brand Fortner and Theodore~E. Meyer.
\newblock {\em Light Spectra to RGB}, pages 47--62.
\newblock Springer New York, New York, NY, 1997.

\bibitem{cie}
John Walker.
\newblock Colour rendering of spectra.
\newblock https://www.fourmilab.ch/documents/specrend/.

\bibitem{Burgot2017}
Jean-Louis Burgot.
\newblock {\em Activities of Electrolytes}, pages 117--133.
\newblock Springer International Publishing, Cham, 2017.

\bibitem{smile}
David Weininger.
\newblock Smiles, a chemical language and information system. 1. introduction to methodology and encoding rules.
\newblock {\em Journal of Chemical Information and Computer Sciences}, 28(1):31--36, 1988.

\bibitem{cas}
Andrea Jacobs, Dustin Williams, Katherine Hickey, Nathan Patrick, Antony~J. Williams, Stuart Chalk, Leah McEwen, Egon Willighagen, Martin Walker, Evan Bolton, Gabriel Sinclair, and Adam Sanford.
\newblock Cas common chemistry in 2021: Expanding access to trusted chemical information for the scientific community.
\newblock {\em Journal of Chemical Information and Modeling}, 62(11):2737--2743, 2022.
\newblock PMID: 35559614.

\bibitem{Iakubovskii}
Pavel Iakubovskii.
\newblock Segmentation models pytorch.
\newblock \url{https://github.com/qubvel/segmentation_models.pytorch}, 2019.

\bibitem{tgcnn}
Shoujie Li, Haixin Yu, Wenbo Ding, Houde Liu, Linqi Ye, Chongkun Xia, Xueqian Wang, and Xiao-Ping Zhang.
\newblock Visual–tactile fusion for transparent object grasping in complex backgrounds.
\newblock {\em IEEE Transactions on Robotics}, 39(5):3838--3856, 2023.

\bibitem{resnet}
Kaiming He, Xiangyu Zhang, Shaoqing Ren, and Jian Sun.
\newblock Deep residual learning for image recognition.
\newblock In {\em Proceedings of the IEEE conference on computer vision and pattern recognition}, pages 770--778, 2016.

\bibitem{unet}
Olaf Ronneberger, Philipp Fischer, and Thomas Brox.
\newblock U-net: Convolutional networks for biomedical image segmentation.
\newblock In {\em Medical image computing and computer-assisted intervention--MICCAI 2015: 18th international conference, Munich, Germany, October 5-9, 2015, proceedings, part III 18}, pages 234--241. Springer, 2015.

\bibitem{VGG}
Karen Simonyan and Andrew Zisserman.
\newblock Very deep convolutional networks for large-scale image recognition.
\newblock {\em arXiv preprint arXiv:1409.1556}, 2014.

\bibitem{unet++}
Zongwei Zhou, Md~Mahfuzur Rahman~Siddiquee, Nima Tajbakhsh, and Jianming Liang.
\newblock Unet++: A nested u-net architecture for medical image segmentation.
\newblock In {\em Deep Learning in Medical Image Analysis and Multimodal Learning for Clinical Decision Support: 4th International Workshop, DLMIA 2018, and 8th International Workshop, ML-CDS 2018, Held in Conjunction with MICCAI 2018, Granada, Spain, September 20, 2018, Proceedings 4}, pages 3--11. Springer, 2018.

\bibitem{efficientnet}
Mingxing Tan and Quoc Le.
\newblock Efficientnet: Rethinking model scaling for convolutional neural networks.
\newblock In {\em International conference on machine learning}, pages 6105--6114. PMLR, 2019.

\bibitem{DeepLabV3}
Liang-Chieh Chen, George Papandreou, Florian Schroff, and Hartwig Adam.
\newblock Rethinking atrous convolution for semantic image segmentation.
\newblock {\em arXiv preprint arXiv:1706.05587}, 2017.

\bibitem{yolo}
Joseph Redmon, Santosh Divvala, Ross Girshick, and Ali Farhadi.
\newblock You only look once: Unified, real-time object detection.
\newblock In {\em Proceedings of the IEEE conference on computer vision and pattern recognition}, pages 779--788, 2016.

\bibitem{PPO}
John Schulman, Filip Wolski, Prafulla Dhariwal, Alec Radford, and Oleg Klimov.
\newblock Proximal policy optimization algorithms.
\newblock {\em arXiv preprint arXiv:1707.06347}, 2017.

\bibitem{gibbs}
Ronald~J. Greaves and Kenneth~D. Schlecht.
\newblock Gibbs free energy: The criteria for spontaneity.
\newblock {\em Journal of Chemical Education}, 69(5):417, 1992.

\end{thebibliography}

}

\end{document}